\documentclass{article}
\PassOptionsToPackage{numbers,compress,sort}{natbib}
\usepackage[preprint]{neurips_2019}
\usepackage{neurips_2019}
\usepackage[utf8]{inputenc}
\usepackage[T1]{fontenc}
\usepackage{amsfonts}
\usepackage{amsmath}
\usepackage{amssymb}
\usepackage{array}
\usepackage{booktabs}
\usepackage{floatrow}
\usepackage{graphicx}
\usepackage{microtype}
\usepackage{nicefrac}
\usepackage{soul}
\usepackage{subcaption}
\usepackage{url}
\usepackage{xcolor}
\usepackage{hyperref}
\usepackage{cleveref}

\newcommand{\cmt}[1]{\ignorespaces}

\title{Photo-geometric autoencoding to learn 3D objects from unlabelled images}
\author{Shangzhe Wu \qquad
\textbf{Christian Rupprecht} \qquad
\textbf{Andrea Vedaldi} \vspace{0.8em} \\
Visual Geometry Group\\
University of Oxford\\
{\tt\small \{szwu, chrisr, vedaldi\}@robots.ox.ac.uk}}

\begin{document}
\maketitle
\begin{abstract}
We show that generative models can be used to capture visual geometry constraints statistically.
We use this fact to infer the 3D shape of object categories from raw single-view images.
Differently from prior work, we use no external supervision, nor do we use multiple views or videos of the objects.
We achieve this by a simple reconstruction task, exploiting the symmetry of the objects' shape and albedo.
Specifically, given a single image of the object seen from an arbitrary viewpoint, our model predicts a symmetric canonical view, the corresponding 3D shape and a viewpoint transformation, and trains with the goal of reconstructing the input view, resembling an auto-encoder.
Our experiments show that this method can recover the 3D shape of human faces, cat faces, and cars from single view images, without supervision.
On benchmarks, we demonstrate superior accuracy compared to other methods that use supervision at the level of 2D image correspondences.
\end{abstract}

\section{Introduction}\label{s:intro}

Given enough labelled data, deep neural networks can learn tasks such as object recognition and monocular depth estimation.
However, doing so from unlabelled data is much more difficult.
In fact, it is often unclear \emph{what} can be learned when labels are missing.
In this paper, we consider the problem of learning the 3D shape of object categories from raw images and seek to solve it by making \emph{minimal} assumptions on the data.
In particular, we do not wish to use any external image annotation.

Given multiple views of the \emph{same} 3D object, techniques such as structure-from-motion (SFM) can be used to reconstruct the object's 3D shape~\cite{Faugeras01}.
In fact, visual geometry suggests that multiple views are not only sufficient, but also necessary for reconstruction.
However, this requirement can be relaxed if one has prior information on the possible shapes of the objects.
By learning such a prior, authors have demonstrated that \emph{monocular} 3D reconstruction is in fact possible and practical~\cite{zhou2017unsupervised, ummenhofer2017demon}.
However, this does not clarify how the necessary prior can be acquired in the first place.
Recent methods such as SFM-Learner~\cite{zhou2017unsupervised} have shown that the prior too can be extracted from a collection of unlabelled images with no externally-provided 3D information.
However, these methods require multiple views of the same object, just as SFM does.

In this paper, we wish to relax these conditions even further and reconstruct objects from an unconstrained collection of images.
By ``unconstrained'', we mean that images are i.i.d.~samples from a distribution of views of different object instances, such as a gallery of human faces.
In particular, each image may contain a different object identity, preventing a direct application of the geometric principles leveraged by SFM and SFM-Learner.
Even so, we argue that these principles are still relevant and applicable, albeit in a statistical sense.

The recent work of~\cite{Moniz2018,Kanazawa18cmr}, and more in general non-rigid SFM approaches, have demonstrated that 3D reconstruction from unconstrained images is possible provided that at least 2D object keypoint annotations are available.
However, the knowledge of keypoints captures a significant amount of information, as shown by the fact that learning a keypoint detector in an unsupervised manner is a very difficult problem in its own right~
\cite{Thewlis17}.
Ultimately, 3D reconstruction of deformable objects without keypoint annotations and from unconstrained images remains an open challenge.

\begin{figure}[t]
\begin{subfigure}{.09\textwidth}
  \includegraphics[width=.99\linewidth]{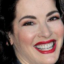}
\end{subfigure}
\hfill
\begin{subfigure}{.8\textwidth}
  \includegraphics[width=.99\linewidth]{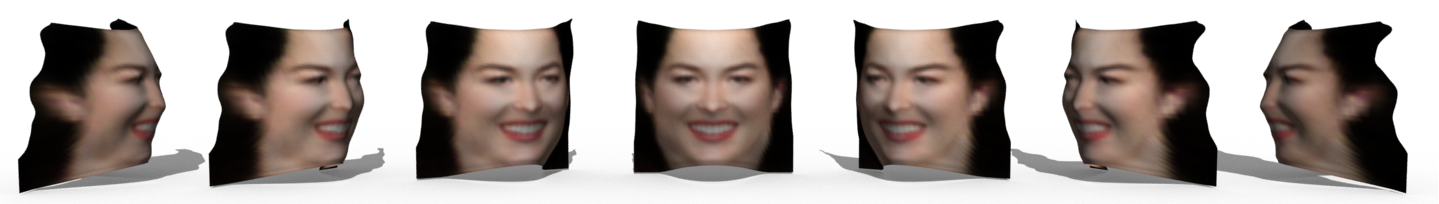}
\end{subfigure} \\

\begin{subfigure}{.09\textwidth}
  \includegraphics[width=.99\linewidth]{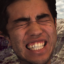}
\end{subfigure}
\hfill
\begin{subfigure}{.8\textwidth}
  \includegraphics[width=.99\linewidth]{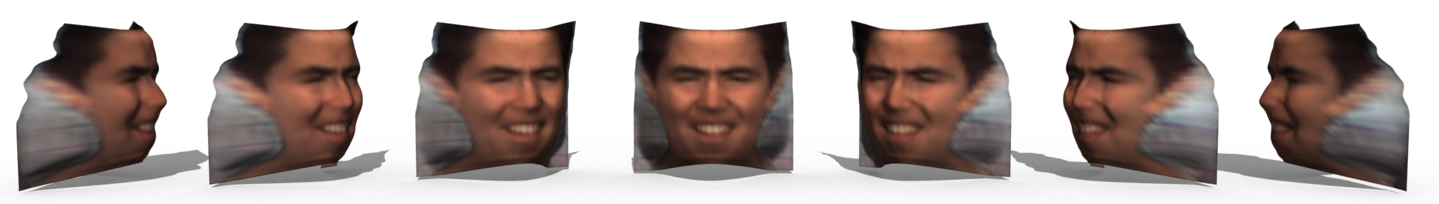}
\end{subfigure} \\ 

\begin{subfigure}{.09\textwidth}
  \includegraphics[width=.99\linewidth]{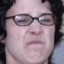}
\end{subfigure}
\hfill
\begin{subfigure}{.8\textwidth}
  \includegraphics[width=.99\linewidth]{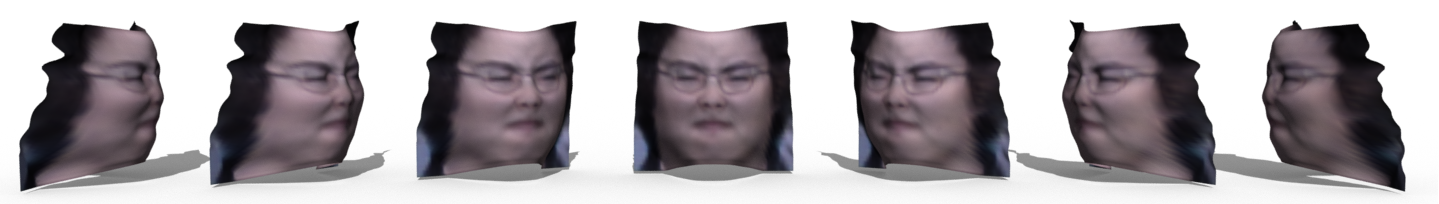}
\end{subfigure} 
\caption{\textbf{From single image to 3D without supervision.} The leftmost column shows the input image to our model and on the right we show the predicted 3D representation from novel viewpoints and modified lighting. During training we only use a collection of ``in-the-wild'' face images with one view per face. Our model learns the geometry, lighting and shading implicitly from the data.}
\label{fig:celeba} 
\end{figure}

In this work, we suggest that the constraints that are explicitly provided by 2D or 3D data annotations can be replaced by weaker  constraints on the statistics of the data that do not require to perform data annotation of any kind.
We capture such constraints as follows (\cref{s:method}).
First, we observe that understanding 3D geometry can explain much of the variation in a dataset of images.
For example, the depth map extracted from an image of a human face depends on the face shape as well as the viewpoint; if the viewpoint is registered, then the variability is significantly reduced.
We can exploit this fact by seeking a generative model that explains the data as a combination of three partially independent factors: viewpoint, geometry and texture.
We pair the generative model with an encoder that can explain any given image of the object as a combination of these factors.
The resulting autoencoder maps the data in a \emph{structured space}.
The structure is specified by how factors are combined:
shape and viewpoint induce, based on the camera geometry, a 2D deformation of the canoncial texture, which in turns matches the observed data.

This process can also be seen as extracting 3D information from 2D image deformations.
However, deformations arising from small depth variations, which are characteristics of objects such as faces, are subtle.
Illumination provides a complementary cue for high-frequency but shallow depth variations.
Using a generative model allows to integrate this cue in a simple manner: we estimate an additional factor, representing the main lighting direction, and combine the latter with the estimated depth and texture to infer the object shading.
Finally, we propose to exploit the fact that many object classes of interest are symmetric, in most cases bilaterally.
We show how this constraint can be leveraged by enforcing that, in the canonical modelling space, appearance and geometry are mirror-symmetric.

We show empirically (\cref{s:exp}) that these cues, when combined, lead to a good monocular reconstruction of 3D object categories such as human and animal faces and objects such as cars.
Quantitatively, we test our model on a large dataset of synthetic faces generated using computer graphics, so that accurate ground-truth information is available for assessment.
We also consider a benchmark dataset of real human faces for which 3D annotations are available and \emph{outperform a recent state-of-the-art method that uses keypoint supervision}, while our method uses no supervision at all.
We also test the components of our methods via an ablation study and show the different benefits they bring to the quality of the reconstruction.

\section{Related Work}\label{s:related_work}

The literature on estimating 3D structures from (collections of) images is vast. Here, we will restrict the overview to mostly learning based methods. 
Traditionally, 3D reconstruction can be achieved by means of multiple view geometry \cite{hartley2003multiple}, where correspondences over multiple views can be used to reconstruct the shape and the camera parameters. 
Another important clue to recover a surface from an image is the shading \cite{woodham1980photometric}. In fact, our model also uses the line between shape and shading to recover the geometry of objects. When multiple distinct views are not available, information can be gained from various different sources.

\cite{zhou2017unsupervised, Wang_2018_CVPR, Novotny17b, Agrawal15} learn from videos, while \cite{monodepth17, Luo2018SVS} train using stereo image pairs. 
Recently, learning instance geometry without multiple views but collections of instance images has emerged. DAE \cite{Shu2018} learns to predict a deformation field through heavily constraining an autoencoder with a small bottleneck embedding. In a recent follow-up work \cite{Sahasrabudhe2019} they learn to disentangle the 3D mesh and the viewpoint from the deformation field. Similarly, SfSNet \cite{Sengupta18sfsnet} learns partially supervised by synthetic ground truth data and \cite{Kanazawa18cmr} needs foreground segmentation and 2D keypoints to learn a parametric 3D model. GANs have been proposed to learn a generative model for 3D shapes by discriminating the reprojection into images \cite{kato2019vpl, Henzler18, Szabo18, nguyen2019hologan, zhu2018von}.
Shape models have been used as a form of supervised constraints to learn a mapping between images and shape parameters \cite{wang2017adversarial, gecer2019ganfit}.
\cite{Thewlis18, Thewlis17b} demonstrate that the symmetry of objects can be leveraged to learn a canonical representation. Depth can also be learned from keypoints \cite{Moniz2018, kudo2018, Chen19, Suwajanakorn2018} which serve as a form of correspondences even across instances of different objects.

Since our model generates images from an internal 3D representation, one part of the model is a differentiable renderer. However, with a traditional rendering pipeline, gradients across occlusions and object boundaries are undefined. Several soft relaxations have thus been proposed \cite{kato2018renderer, Liu2018softras, Loper2014}. In this work we use an implementation\footnote{\url{https://github.com/daniilidis-group/neural_renderer}} of \cite{kato2018renderer}.

\section{Method}\label{s:method}

\begin{figure}[t]
\begin{subfigure}{.6\textwidth}
  \includegraphics[height=15em]{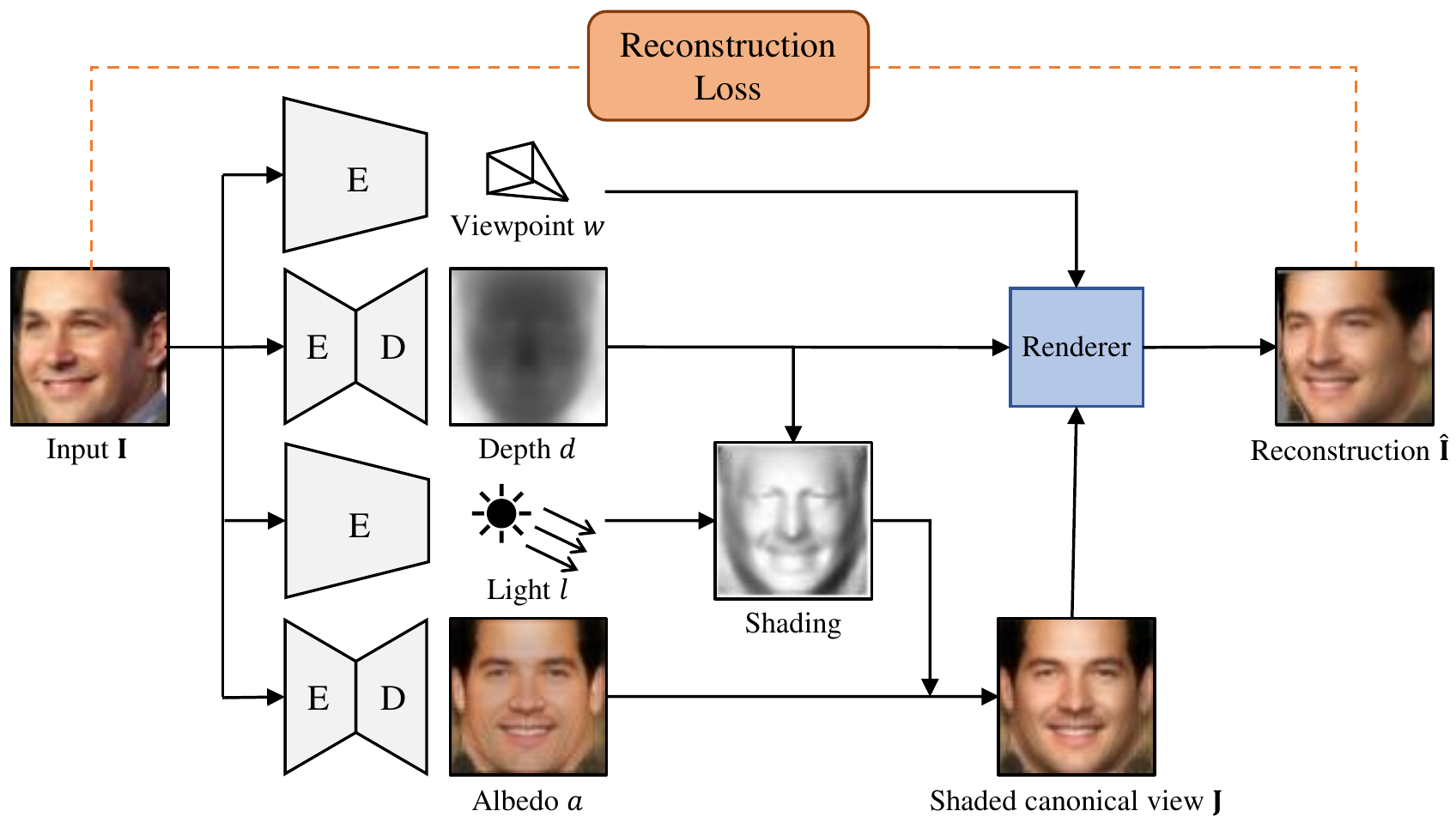}
  \caption{Unsupervised model.}\label{fig:pipeline}
\end{subfigure}
\hfill
\begin{subfigure}{.3\textwidth}
  \includegraphics[height=15em,trim=0 0 0 -8em]{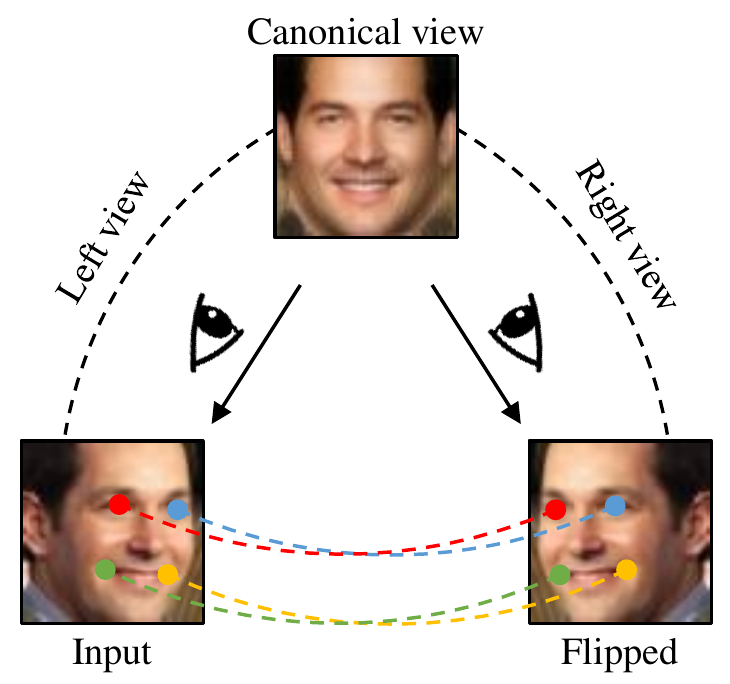}
  \caption{Canonical view and symmetry.}\label{fig:symmetry}
\end{subfigure}
\caption{
Left: Our network $\Phi$ decomposes an input image $\mathbf{I}$ into shape, albedo, viewpoint and shading. It is trained in an unsupervised fashion to reconstruct the input images.
Right: The process is regularized by mapping the image into a space where the symmetry of the object is apparent.}
\end{figure}

Our method takes as input an unconstrained collection of images of an object category, such as human faces, and returns as output a model $\Phi$ that can explain each image as the combination of a 3D shape, a texture, an illumination and a viewpoint, as illustrated in~\cref{fig:pipeline}.

Formally, an image $\mathbf{I}$ is a function $\Omega\rightarrow\mathbb{R}^3$ defined on a lattice $\Omega = \{0, \ldots, W-1\} \times \{0, \ldots, H-1\}$, or, equivalently, a tensor in $\mathbb{R}^{3\times W\times H}$.
We assume that the image is roughly centered on an instance of the objects of interest.
The goal is to learn a function $\Phi$, implemented as a neural network, that maps the image $\mathbf{I}$ to four factors $(d, a, w, l)$ consisting of a \emph{depth map} $d : \Omega \rightarrow \mathbb{R}_+$, and \emph{albedo image} $a : \Omega \rightarrow \mathbb{R}^3$, a \emph{viewpoint} $w \in \mathbb{R}^6$ and a global \emph{light direction} $l \in \mathbb{S}^2$.

In order to learn this disentangled representation without supervision, we task the model $\Phi$ with the goal of reconstructing the input image $\mathbf{I}$ from the four factors.
The reconstruction is a differentiable operation composed of two steps: \emph{lighting} $\Lambda$ and \emph{reprojection} $\Pi$, as follows:
\begin{equation}\label{e:generator}
\hat{\mathbf{I}} = \Pi\left(\Lambda(a, d, l), d, w\right).
\end{equation}
The lighting function $\Lambda$ generates a version of the face based on the depth map $d$, the light direction $l$ and the albedo $a$ as seen from a canonical viewpoint $w=0$.
For example, for faces a natural choice for the canonical viewpoint is a frontal view, since this minimizes self occlusions, but we let the network choose one automatically.
The viewpoint $w$ then represents the transformation between the canonical view and the viewpoint of the actual input image $\mathbf{I}$.
Then, the reprojection function $\Pi$ simulates the effect of a viewpoint change and generates the image $\mathbf{I}$ given the canonical depth $d$ and the shaded canonical image $\Lambda(a, d, l)$.

Next, we discuss the functions $\Pi$ and $\Lambda$ in model~\eqref{e:generator} in detail.

\paragraph{Reprojection function $\Pi$ and camera model.}

The image is formed by a camera sensor looking at a 3D object.
If we denote with $P = (P_x,P_y,P_z)\in\mathbb{R}^3$ a 3D point expressed in the reference frame of the camera, this is mapped to pixel
$p = (u,v,1)$ by the following projection equation:
\begin{equation}\label{e:camera}
p \propto K P
\quad\text{where}\quad
K =
\begin{bmatrix}
f & 0 & c_u \\
0 & f & c_v \\
0 & 0 & 1 \\
\end{bmatrix}
\quad\text{and}\quad
\begin{cases}
c_u=\frac{W-1}{2},\\
c_v=\frac{H-1}{2},\\
f = \frac{W}{2\tan\frac{\theta_{\text{FOV}}}{2}}.\\
\end{cases}
\end{equation}
This model assumes a perspective camera with \emph{field of view} (FOV) $\theta_\text{FOV}$ in the horizontal direction.
Given that the images are cropped around a particular object, we assume a relatively narrow FOV of $\theta_{\text{FOV}} \approx 25^\circ$.
The object is assumed to be approximately at a distance of $0.5\mathrm{m}$ from the camera.

The depth map $d : \Omega\rightarrow\mathbb{R}_+$ associates to each pixel $(u,v) \in \Omega$ a depth value $d_{uv}$ in the canonical view.
Inverting the camera model~\eqref{e:camera}, this corresponds to the 3D point
$
 P = d_{uv} \cdot K^{-1} p.
$


The viewpoint $w$ represents an Euclidean transformation $(R,T)\in SE(3)$ such that $R=\exp \hat w_{1:3}$ (this is the exponential map of $SO(3)$) and $T = w_{4:6}$.

The map $(R,T)$ transforms 3D points from the canonical view to the actual view.
Thus a pixel $(u,v)$ in the canonical view is mapped to the pixel $(u', v')$ in the actual view by the warping function $(u',v') = \eta_{d,w}(u,v)$ given by:
\begin{equation}\label{e:forward}
p'  \propto K (d_{uv} \cdot R K^{-1} p + T),
~\text{where}~
p' = \begin{bmatrix} u' \\ v' \\ 1 \end{bmatrix}
~\text{and}~
p = \begin{bmatrix} u \\ v\\ 1 \end{bmatrix}.
\end{equation}
Finally, the reprojection function $\Pi$ takes as input the depth $d$ and the viewpoint change $w$ and applies the resulting warp to the canonical image  $\mathbf{J}$ to obtain the actual image $\mathbf{I} = \Pi(\mathbf{J},d,w)$ as
$$
  \mathbf{I}_{u'v'} = \mathbf{J}_{uv},
  \quad
  \text{where}
  \quad
  (u, v) = \eta_{d,w}^{-1}(u', v').
$$
Notice that this requires to compute the \emph{inverse} of the warp $\eta_{d,w}$.
This issue is discussed in detail in~\cref{sec:render}.

\paragraph{Lighting function $\Lambda$.}\label{sec:shading}

The goal of the lighting function $\mathbf{J} = \Lambda(a, d, l)$ is to generate the canonical image as a combination of albedo, 3D shape and light direction.
Note that the effect of lighting could be incorporated in the factor $a$ by interpreting the latter as a texture rather than as the object's albedo.
However, there are two good reasons for avoiding this.
First, the albedo $a$ is often symmetric even if the illumination causes the corresponding texture to look asymmetric.
Separating them allows us to more effectively incorporate the symmetry constraint described below.
Second, shading provides an additional cue on the underlying 3D shape~\cite{woodham1980photometric, horn75obtaining, belhumeur99thebasrelief}.
In particular, unlike the recent work of~\cite{Shu2018} where a shading map is predicted independently from shape, our model computes the shading based on the predicted depth, constraining the two.

Formally, given the depth map $d$, we derive the normal map
$n : \Omega \rightarrow \mathbb{S}^2$ by associating to each pixel $(u,v)$ a vector normal to the underlying 3D surface.
In order to find this vector, we compute the vectors  $t^u_{uv}$ and $ t^v_{uv}$ tangent to the surface along the $u$ and $v$ directions.
For example, the first one is:
$$
t^u_{uv} =
d_{u+1,v}\cdot
K^{-1}
\begin{bmatrix}
u+1 \\ v \\ 1
\end{bmatrix}-
d_{u-1,v}
\cdot
K^{-1}
\begin{bmatrix}
u-1 \\ v \\ 1
\end{bmatrix}.
$$
Then the normal is obtained by taking the vector product $n_{uv} \propto t^u_{uv} \times t^v_{uv}$.

The normal $n_{uv}$ is multiplied by the light direction $l$ to obtain a value for the direct illumination and the latter is added to the ambient light.
Finally, the result is multiplied by the albedo function to obtain the illuminated texture, as follows:
\begin{equation}\label{e:illumination}
\forall (u,v) \in \Omega:
\qquad
 \mathbf{J}_{uv} =
 \left(k_s + k_d \max\{0, \langle l , n_{uv} \rangle\} \right) \cdot a_{uv}.
\end{equation}
Here $k_s$ and $k_d$ are the scalar coefficients weighting the ambient and direct terms.

\subsection{Differentiable rendering layer}\label{sec:render}

As noted in the previous section, the reprojection function $\Pi$ \emph{warps} the canonical image $\mathbf{J}$ to generate the actual image $\mathbf{I}$.
In CNNs, image warping is usually regarded as a simple operation that can be implemented efficiently using a bilinear resampling layer~\cite{jaderberg2015spatial}.
However, this is true only if we can easily send pixels $(u',v')$ in the warped image $\mathbf{I}$ back to pixels $(u,v)$ in the source image $\mathbf{J}$, a process also known as \emph{backward warping}.
Unfortunately, in our case the function $\eta_{d,w}$ obtained by~\cref{e:forward} sends pixels in the opposite direction.

Implementing a \emph{forward warping} layer is surprisingly delicate.
One way of approaching the problem is to regard this task as a special case of rendering a textured mesh.
The recent \emph{Neural Mesh Renderer} (NMR) of~\cite{kato2018renderer} is a differentiable renderer of this type.
In our case, however, the mesh has one vertex per pixel and each group of $2\times 2$ adjacent pixels is tessellated by two triangles.
Empirically, we found the quality of the texture gradients computed by NMR to be poor in this case, probably also due to the high frequency content of the texture image $\mathbf{J}$.

We solve the problem as follows.
First, we use NMR to warp not the albedo $a$, but the depth map $d$ itself, obtaining a version $d'$ of the depth map as seen from the actual viewpoint.
This has two advantages: NMR is much faster when the task is limited to rendering the depth map instead of warping an actual texture.
Secondly, the gradients are more stable, probably also due to the comparatively smooth nature of the depth map $d$ compared to the texture image $\mathbf{J}$.
Given the depth map $d'$, we then use the inverse of~\eqref{e:forward} to find the warp field from the observed viewpoint to the canonical viewpoint, and bilinearly resample the canonical image $\mathbf{J}$ to obtain the reconstruction.


\paragraph{Discussion.}

Several alternative architectures were tested and discarded in favor of the one outlined above.
Among those, one option is to task the network to estimate $d$ as well as $d'$.
However, this requires to ensure that the two depth maps are compatible, which adds extra complexity to the model and did not work as well.

\subsection{Symmetry}


A constraint that is often useful in modelling object categories is the fact that these have a \emph{bilateral symmetry}, both in shape and albedo.
Under the assumption of bilateral symmetry, we are able to obtain a second virtual view of an object simply by flipping the image horizontally, as shown in \cref{fig:symmetry}.
Note that, if we are given the correspondence between symmetric points of the object (such as the corner of the two eyes, etc.), we could use this information to infer the object's 3D shape~\cite{gao17exploiting, gordon90shape}.
While such correspondences are not given to us as the system is unsupervised, we estimate them implicitly by mapping the image to the canonical space.

In practice, there are various ways to enforce a symmetry constraint.
For example, one can add a symmetry loss term to the learning objective as a regularizer.
However, this requires balancing more terms in the objective.
Instead, we incorporate symmetry by performing reconstruction from both canonical image and its mirrored version.

In order to do so, we introduce the (horizontal) flipping operator, whose action on a tensor $a\in\mathbb{R}^{C\times W\times H}$ is given by
$
  [\operatorname{flip} a]_{c,u,v} = a_{c,W-1-u,v}.
$
During training, we randomly choose to flip the canonical albedo $a$ and depth $d$ before we reconstruct the image using~\cref{e:generator}.
Implicitly and without introducing an additional loss term, this imposes several constraints on the model.
Both, depth and albedo will be predicted with horizontal symmetry by $\Phi$ to overcome the confusion that is introduced by the flipping operation.
Additionally, this constrains the canonical viewpoint $w_0$ to align the object's plane of symmetry with the vertical centerline of the image.
Finally, flipping helps to disentangle albedo and shading:
if an object is lit from one side and the albedo is flipped, the target still needs to be lit from the same side, requiring the shading to arise from geometry and not from the texture.

\subsection{Loss, regularizer, and objective function}


The primary loss function of our model is the $\ell_1$ loss on the reconstruction and input image $\mathbf{I}$:
\begin{equation}\label{e:l1}
\mathcal{L}_1(\mathbf{I};\Phi) = \|\hat{\mathbf{I}} - \mathbf{I} \|_1,
\qquad
\hat{\mathbf{I}} = \Pi\left(\Lambda(a, d, l), d, w\right),
\qquad
(d, a, w, l) = \Phi(\mathbf{I}).
\end{equation}
However, this loss is sensitive to small geometric imperfection and tends to result in blurry reconstruction; to avoid that, we add a \emph{perceptual loss}, which is more robust to such geometric imperfections and eventually leads to a much sharper canonical image.
This is obtained by using an off-the-shelf image encoder $e$ (VGG16 in our case~\cite{Simonyan15}), and is given by
  $
\mathcal{L}_{\text{perc}}(\mathbf{I};\Phi)
=
\sum_{k=1}^K \frac{1}{C_k H_k W_k}
\|
e_k(\hat{\mathbf{I}}) -
e_k(\mathbf{I})
\|^2_F
$
where $e_k(I) \in \mathbb{R}^{C_k\times H_k\times W_k}$ is the feature map computed by the $k$-th layer of the encoder network.


We regularize the viewpoint by pulling its mean to zero, breaking the tie between equivalent rotations (which have a period of $360^{\circ}$) and aligning the canonical view $w_0$ to the mean viewpoint in the dataset.
This is achieved by minimizing the function
$
\mathcal{R}_\text{vp}(\Phi;\mathcal{B}) = 
\|
\frac{1}{B} \sum_{i=1}^B w^{(i)}
\|_2^2
$
where $w^{(i)}$ is the viewpoint is estimated for image $\mathbf{I}_i$ in a batch $\mathcal{B}=\{\mathbf{I}_i, i=1,\dots,B\}$ of $B$ images.
%
We also regularize the depth by shrinking its variance between faces.
We do so via the regularization term
$
\mathcal{R}_{\text{d}}(\Phi;\mathbf{I}_i,\mathbf{I}_j) =
\| d^{(i)} - d^{(j)} \|^2_F
$
where $d^{(i)}$ and $d^{(j)}$ are the depth maps obtained from a \emph{pair} of example images $\mathbf{I}_i$ and $\mathbf{I}_j$.
%
%
Losses and regularizers are averaged over a batch, yielding the objective:
\begin{equation}
  \mathcal{E}(\Phi;\mathcal{B}) =
  \frac{1}{B}
  \sum_{i=1}^B
  \left[
  \lambda_1 \mathcal{L}_1(\Phi;\mathbf{I}_i) +
  \lambda_\text{perc} \mathcal{L}_\text{perc}(\Phi;\mathbf{I}_i) +
  \frac{2\lambda_\text{d}}{B-1}
  \sum_{j=i+1}^B
  \mathcal{R}_\text{d}(\Phi;\mathbf{I}_i,\mathbf{I_j})
  \right] + \lambda_\text{vp} \mathcal{R}_\text{vp}(\Phi;\mathcal{B}).
\end{equation}

\subsection{Neural network architecture}\label{s:arch}

We use different networks to extract depth, albedo, viewpoint and lighting $(d, a, w, l) = \Phi(\mathbf{I})$ from a single image $\mathbf{I}$ of the object.
The depth and albedo are generated by encoder-decoder networks, while viewpoint and lighting are regressed using simple encoder networks.
In particular, we use DenseNet~\cite{huang2017densely} for albedo prediction, with deeper architecture than standard encoder-decoder for depth prediction, because we would like the albedo to capture more details than the depth.
We do not use skip connections between encoder and decoder because the network is generating a different view, and thus pixel alignment is not desirable.

\section{Experiments}\label{s:exp}

\begin{table}[t]
\newcommand{\xpm}[1]{{\tiny$\pm#1$}}
\setlength{\tabcolsep}{1pt}
\begin{floatrow}[2]
  \ttabbox{
    \footnotesize
    \begin{tabular}{clp{7em}p{6em}}  
      \toprule
      No.&Method                      & Scale-inv.~err.~($\times10^{-2}$) & Normal err.~(deg.)\\\midrule
      (1)&Supervised ($\ell_1$)       & $1.368$\xpm{0.288}                & $11.290$\xpm{1.188}\\
      (2)&Supervised ($\ell_2$)       & $1.437$\xpm{0.263}                & $13.242$\xpm{1.284}\\\midrule
      (3)&Const.~null depth           & $7.839$\xpm{0.742}                & $41.604$\xpm{2.873}\\
      (4)&Const.~blob depth           & $5.687$\xpm{0.879}                & $27.152$\xpm{2.226}\\
      (5)&Average g.t.~depth          & $5.029$\xpm{1.267}                & $23.963$\xpm{3.454}\\ \midrule
      (6)&Ours                        & $3.143$\xpm{0.592}                & $17.650$\xpm{1.873}\\
    \bottomrule
    \end{tabular}
  }{
    \caption{\textbf{Performance bounds.} We compare our method to fully-supervised and trivial baselines.}\label{tab:syn_numbers}
  }
  \ttabbox{
    \footnotesize
    \begin{tabular}{lr}
      \toprule
                                                         & Depth Corr. \\ \midrule
      Ground truth                                       & $66$ \\
      AIGN \cite{tung2017} (\textbf{supervised})         & $50.81$ \\
      DepthNetGAN \cite{Moniz2018} (\textbf{supervised}) & $58.67$ \\ \midrule
      MOFA \cite{tewari17MoFA} (model-based)             & $15.97$ \\
      DepthNet (paper) \cite{Moniz2018}                  & $26.32$ \\
      DepthNet (github) \cite{Moniz2018}                 & $35.77$ \\ \midrule 
      Ours & $51.65$ \\
      \bottomrule 
    \end{tabular}
    }{\caption{\textbf{3DFAW - Keypoint depth.} Depth correlation between ground truth and prediction evaluated at facial keypoint locations.} \label{tab:3dfaw_kpt}}
\end{floatrow}
\end{table}


We first analyze the contribution of the individual components of our model~\eqref{e:generator} and of the regularizers.
We do so quantitatively, by using a synthetic face dataset where 3D ground truth is available to measure the quality of the predicted depth maps.
However, we also show qualitatively that these metrics do not fully account for all the aspects that make a good 3D reconstruction, and we demonstrate that some components of our model are particularly good at addressing those.
We also compare our method on real data to~\cite{Moniz2018} who estimate depth for facial keypoints, and test the generalization of the model to objects other than human faces by training on synthetic ShapeNet cars and real cat faces. 

For reproducibility and future comparisons, we describe network architecture details and hyperparameters in the supplementary material.
We will release the code, trained models and the synthetic dataset upon acceptance of the paper. 

\subsection{Quantitative assessment and ablation}

To evaluate the model quantitatively, we utilize synthetic data, where we know the ground truth depth.
We follow in particular the protocol of~\cite{Sengupta18sfsnet} to generate a large dataset of synthetic faces using the Basel Face Model~\cite{bfm09}.
The faces are rendered with shapes, textures, illuminations, and rotations randomly sampled from the model.
We use images from SUN Database~\cite{Xiao2010SUN} as background and render the images together with ground truth depth maps for evaluation. 

Since the scale of 3D reconstruction from projective cameras is inherently ambiguous, we adjust for it in the evaluation.
Specifically, we take the depth map $d$ predicted by the model in the canonical view, map it to a depth map $d'$ in the actual view, and compare the latter to the ground-truth depth map $d^*$ using the scale invariant error~\cite{Eigen2014}
$
    E_\text{SI}(d, d^*) = 
        (
            \frac{1}{WH} \sum_{u,v}\Delta_{uv}^2 -
            (
                \frac{1}{WH}\sum_{u,v}\Delta_{uv}
            )^2
        )^{\frac{1}{2}}.
$
where $\Delta_{uv} = \log d_{uv} - \log d^*_{uv}$.
Additionally, we report the mean angle deviation between normals computed from ground truth depth and from the depth prediction, which measures how well the surface is captured by the prediction.

In~\cref{tab:syn_numbers} we estimate upper bounds on the model performance by comparing it to supervised baselines using the same network architectures.
We also see a significant improvement over various constant prediction baselines.

\begin{figure}[t]
  \includegraphics[width=.9\linewidth,trim=0 6.5cm 0 0]{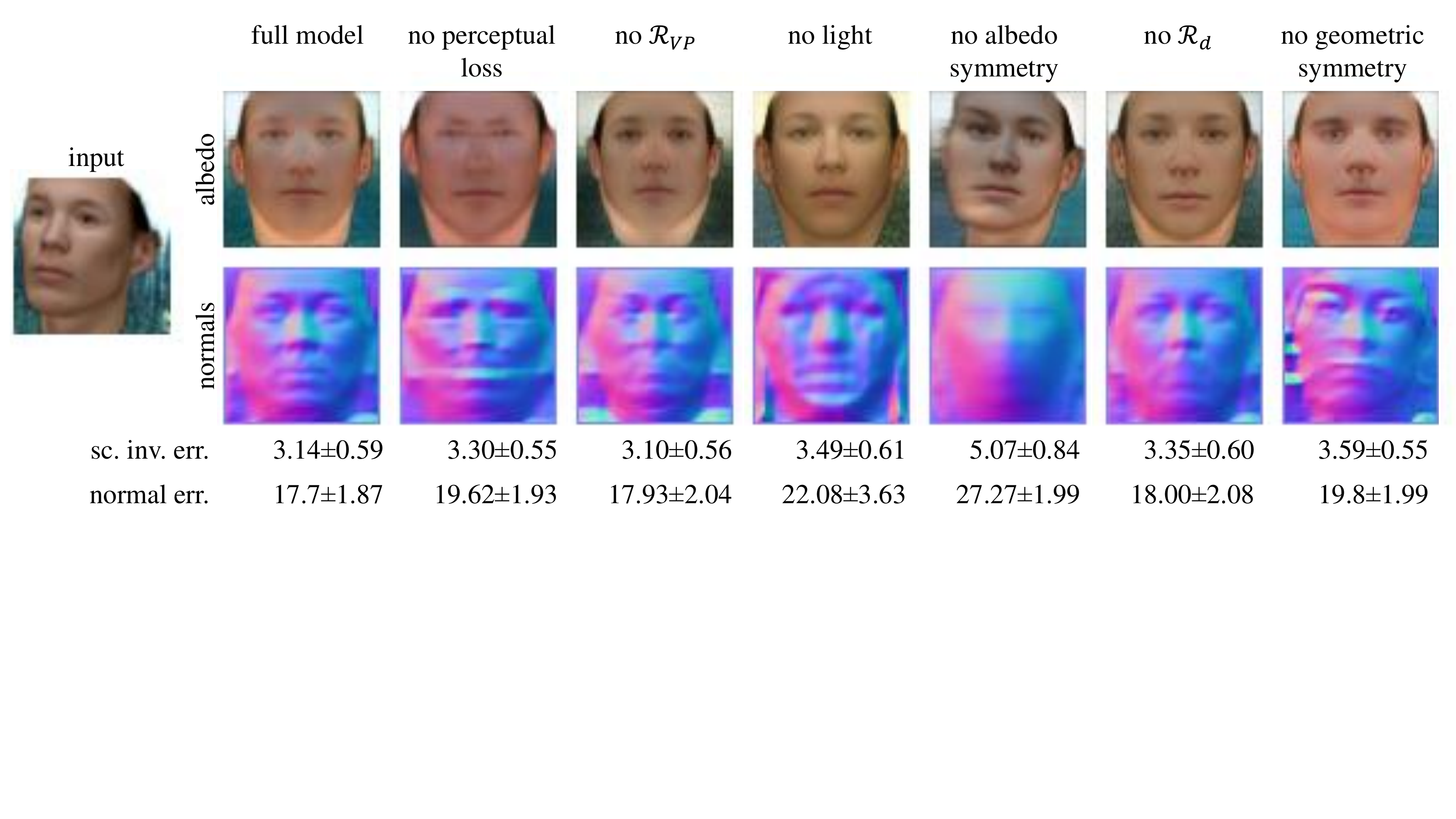}
  \caption{\textbf{Ablation study.} We compute the overall test set performance when turning off different components of our model and report the scale invariant error and absolute angle error (in degrees).
  Additionally, we show for one image, the canonical albedo $a$ and normal map computed from canonical depth $d$.}
  \label{fig:ablation}
\end{figure}

To understand the influence of the individual parts, we remove each one of them and evaluate the ablated model in \cref{fig:ablation}.
However, since the error $E_\text{SI}$ is computed in the original view of the image, it does not evaluate the quality of the 3D shape not visible from that vantage point.
Thus, we also visualize the canonical albedo $a$ and the normal map computed from the depth map $d$.
We can see that all components reduce the error as well as improve the visual quality of the samples.
The symmetry constraint for the albedo and depth have the strongest impact on the model, while the perceptual loss improves the quality of the reconstruction and helps to avoid local minima during training.
The regularizers improve the canonical representation, and, lastly, lighting helps to resolve possible ambiguities in the geometric reconstruction, particularly in texture-less regions.

\subsection{Qualitative results}

To evaluate the performance on real data, we conduct experiments using two datasets.
CelebA~\cite{liu2015celeba} contains over $200$k images of real human faces, and 3DFAW~\cite{multipie2010, Jeni2015, ZHANG2014692, Yin2008} contains $23$k images with $66$ 3D keypoint annotations.
We crop the faces from original images using the provided keypoints, and follow the official train/val/test splits.
In~\cref{fig:celeba} we show qualitative results on both datasets.
The 3D shape of the faces is recovered very well by the model, including details of nose, eyes and mouth, despite the presence of extreme facial expression. 

\subsection{Comparison with the state of the art} \label{s:exp_sota}

To the best of our knowledge, there is no prior work on fully unsupervised dense object depth estimation that we can directly compare with. 
However, the DepthNet model of~\cite{Moniz2018} predicts depth for selected facial keypoints given the 2D keypoint locations as input.
Hence, we can evaluate the reconstruction obtained by our method on this sparse set of points.
We also compare to the baselines MOFA~\cite{tewari17MoFA} and AIGN~\cite{tung2017} reported in~\cite{Moniz2018}.
For a fair comparison, we use their public code which computes the depth correlation score (between $0$ and $66$).
We use the 2D keypoint locations to sample our predicted depth and then evaluate the same metric.
The set of test images from 3DFAW~\cite{multipie2010, Jeni2015, ZHANG2014692, Yin2008} and the preprocessing are identical to \cite{Moniz2018}. 

In~\cref{tab:3dfaw_kpt} we report the results from their paper and the slightly higher results we obtained from their publicly-available implementation.
The paper also evaluates a supervised model using a GAN discriminator trained with ground-truth depth information.
Our fully unsupervised model outperforms DepthNet and reaches close-to-supervised performance, indicating that we learn reliable depth maps. 

\subsection{Generalization to other objects}

To understand the generalization of the method to other symmetric objects, we train on two additional datasets.
We use the cat dataset provided by~\cite{zhang2008cat}, crop the cat heads using the  keypoint annotations and split the images by $8$:$1$:$1$ into train, validation and test sets.
For car images we render ShapeNet's~\cite{shapenet2015} synthetic car models from various viewpoints and textures. 

\begin{figure}[t]
\begin{subfigure}{.09\textwidth}
  \includegraphics[width=.99\linewidth]{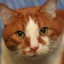}
\end{subfigure}
\hfill
\begin{subfigure}{.80\textwidth}
  \includegraphics[width=.99\linewidth]{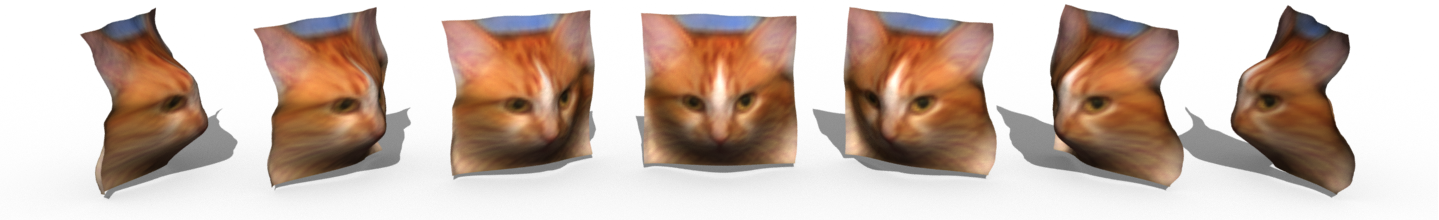}
\end{subfigure} \\ 

\begin{subfigure}{.09\textwidth}
  \includegraphics[width=.99\linewidth]{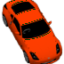}
\end{subfigure}
\hfill
\begin{subfigure}{.81\textwidth}
  \includegraphics[width=.99\linewidth]{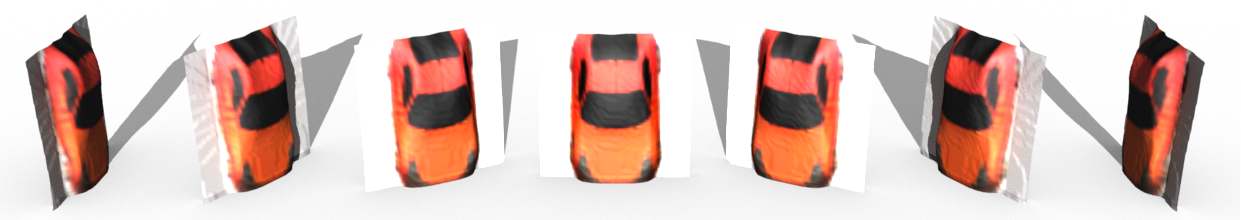}
\end{subfigure} 
\caption{\textbf{Other datasets.} Results from training on cat faces and cars.} \label{fig:catscars}
\end{figure}

We are able to reconstruct both object categories well and the results are visualized in \cref{fig:catscars}.
Although we assume Lambertian surfaces to estimate the shading, our model can reconstruct cat faces convincingly despite their fur which has complicated light transport mechanics.
This shows that the other parts of the model constrain the shape enough to still converge to meaningful representations. 
Overall, the model is able to reconstruct cats and cars as well as human faces, showing that the method generalizes over object categories.
\section{Conclusions}\label{s:conc}

We have presented a method that can learn from an unconstrained image collection of single views of a given category to reconstruct the 3D shapes of individual instances.
The model is fully unsupervised and learns based on a reconstruction loss, similar to an autoencoder.
We have shown that lighting and symmetry are strong indicators for shape and help the model to converge to a meaningful reconstruction.
Our model outperforms a current state-of-the-art method that uses 2D keypoint supervision.
As for future work, the model currently represents 3D shape from a canonical viewpoint, which is sufficient for objects such as faces that have roughly convex shape and a natural canonical viewpoint.
In order to handle more complex objects, it may be possible to extend the model to use either a collection of canonical views or a 3D representations such as a mesh or a voxel map.
\paragraph{Acknowledgement}
We gratefully thank Soumyadip Sengupta for sharing with us the code to generate synthetic face datasets, and members of Visual Geometry Group for insightful discussion. Shangzhe Wu is supported by Facebook Research. Christian Rupprecht is supported by ERC Stg Grant IDIU-638009. 

\bibliography{shortstrings,vgg_local,vgg_other,refs}

\begin{thebibliography}{50}
\providecommand{\natexlab}[1]{#1}
\providecommand{\url}[1]{\texttt{#1}}
\expandafter\ifx\csname urlstyle\endcsname\relax
  \providecommand{\doi}[1]{doi: #1}\else
  \providecommand{\doi}{doi: \begingroup \urlstyle{rm}\Url}\fi

\bibitem[Agrawal et~al.(2015)Agrawal, Carreira, and Malik]{Agrawal15}
Pulkit Agrawal, Joao Carreira, and Jitendra Malik.
\newblock Learning to see by moving.
\newblock In \emph{Proc. ICCV}, pages 37--45. IEEE, 2015.

\bibitem[Belhumeur et~al.(1999)Belhumeur, Kriegman, and
  Yuille]{belhumeur99thebasrelief}
P.~N. Belhumeur, D.~J. Kriegman, and A.~L. Yuille.
\newblock The bas-relief ambiguity.
\newblock \emph{IJCV}, 35\penalty0 (1), 1999.

\bibitem[Chang et~al.(2015)Chang, Funkhouser, Guibas, Hanrahan, Huang, Li,
  Savarese, Savva, Song, Su, Xiao, Yi, and Yu]{shapenet2015}
A.~X. Chang, T.~Funkhouser, L.~Guibas, P.~Hanrahan, Q.~Huang, Z.~Li,
  S.~Savarese, M.~Savva, S.~Song, H.~Su, J.~Xiao, L.~Yi, and F.~Yu.
\newblock Shapenet: An information-rich 3d model repository.
\newblock \emph{arXiv}, abs/1512.03012, 2015.

\bibitem[Chen et~al.(2019)Chen, Tyagi, Agrawal, Drover, Rohith, Stojanov, and
  Rehg]{Chen19}
C.{-}H. Chen, A.~Tyagi, A.~Agrawal, D.~Drover, M.~V. Rohith, S.~Stojanov, and
  J.~M. Rehg.
\newblock Unsupervised 3d pose estimation with geometric self-supervision.
\newblock \emph{arXiv}, abs/1904.04812, 2019.

\bibitem[Eigen et~al.(2014)Eigen, Puhrsch, and Fergus]{Eigen2014}
D.~Eigen, C.~Puhrsch, and R.~Fergus.
\newblock Depth map prediction from a single image using a multi-scale deep
  network.
\newblock In \emph{NeurIPS}, 2014.

\bibitem[Faugeras and Luong(2001)]{Faugeras01}
O.~Faugeras and Q.-T. Luong.
\newblock \emph{The Geometry of Multiple Images}.
\newblock MIT Press, 2001.

\bibitem[Gao and Yuille(2017)]{gao17exploiting}
Y.~Gao and A.~L. Yuille.
\newblock Exploiting symmetry and/or manhattan properties for 3d object
  structure estimation from single and multiple images.
\newblock In \emph{Proc. CVPR}, 2017.

\bibitem[Gecer et~al.(2019)Gecer, Ploumpis, Kotsia, and
  Zafeiriou]{gecer2019ganfit}
B.~Gecer, S.~Ploumpis, I.~Kotsia, and S.~Zafeiriou.
\newblock Ganfit: Generative adversarial network fitting for high fidelity 3d
  face reconstruction.
\newblock \emph{arXiv}, abs/1902.05978, 2019.

\bibitem[Godard et~al.(2017)Godard, Mac~Aodha, and Brostow]{monodepth17}
C.~Godard, O.~Mac~Aodha, and G.~J. Brostow.
\newblock Unsupervised monocular depth estimation with left-right consistency.
\newblock In \emph{Proc. CVPR}, 2017.

\bibitem[Gordon(1990)]{gordon90shape}
G.~G. Gordon.
\newblock Shape from symmetry.
\newblock In \emph{Proc. {SPIE}}, 1990.

\bibitem[Gross et~al.(2010)Gross, Matthews, Cohn, Kanade, and
  Baker]{multipie2010}
R.~Gross, I.~Matthews, J.~Cohn, T.~Kanade, and S.~Baker.
\newblock Multi-pie.
\newblock \emph{Image and Vision Computing}, 2010.

\bibitem[Hartley and Zisserman(2003)]{hartley2003multiple}
R.~Hartley and A.~Zisserman.
\newblock \emph{Multiple view geometry in computer vision}.
\newblock Cambridge university press, 2003.

\bibitem[Henzler et~al.(2018)Henzler, Mitra, and Ritschel]{Henzler18}
P.~Henzler, N.~J. Mitra, and T.~Ritschel.
\newblock Escaping plato's cave using adversarial training: 3d shape from
  unstructured 2d image collections.
\newblock \emph{arXiv}, abs/1811.11606, 2018.

\bibitem[Horn(1975)]{horn75obtaining}
B.~Horn.
\newblock Obtaining shape from shading information.
\newblock In \emph{The Psychology of Computer Vision}, 1975.

\bibitem[Huang et~al.(2017)Huang, Liu, van~der Maaten, and
  Weinberger]{huang2017densely}
G.~Huang, Z.~Liu, L.~van~der Maaten, and K.~Q. Weinberger.
\newblock Densely connected convolutional networks.
\newblock In \emph{Proc. CVPR}, 2017.

\bibitem[Jaderberg et~al.(2015)Jaderberg, Simonyan, Zisserman, and
  Kavukcuoglu]{jaderberg2015spatial}
M.~Jaderberg, K.~Simonyan, A.~Zisserman, and K.~Kavukcuoglu.
\newblock Spatial transformer networks.
\newblock In \emph{NeurIPS}, 2015.

\bibitem[Jeni et~al.(2015)Jeni, Cohn, and Kanade]{Jeni2015}
L.~A. Jeni, J.~F. Cohn, and T.~Kanade.
\newblock Dense 3d face alignment from 2d videos in real-time.
\newblock In \emph{Proc. Int. Conf. Autom. Face and Gesture Recog.}, 2015.

\bibitem[Kanazawa et~al.(2018)Kanazawa, Tulsiani, Efros, and
  Malik]{Kanazawa18cmr}
A.~Kanazawa, S.~Tulsiani, Alexei~A. Efros, and J.~Malik.
\newblock Learning category-specific mesh reconstruction from image
  collections.
\newblock In \emph{Proc. ECCV}, 2018.

\bibitem[Kato and Harada(2019)]{kato2019vpl}
H.~Kato and T.~Harada.
\newblock Learning view priors for single-view 3d reconstruction.
\newblock In \emph{Proc. CVPR}, 2019.

\bibitem[Kato et~al.(2018)Kato, Ushiku, and Harada]{kato2018renderer}
H.~Kato, Y.~Ushiku, and T.~Harada.
\newblock Neural 3d mesh renderer.
\newblock In \emph{Proc. CVPR}, 2018.

\bibitem[Kudo et~al.(2018)Kudo, Ogaki, Matsui, and Odagiri]{kudo2018}
Y.~Kudo, K.~Ogaki, Y.~Matsui, and Y.~Odagiri.
\newblock Unsupervised adversarial learning of 3d human pose from 2d joint
  locations.
\newblock \emph{arXiv}, abs/1803.08244, 2018.

\bibitem[Liu et~al.(2019)Liu, Li, Chen, and Li]{Liu2018softras}
S.~Liu, T.~Li, W.~Chen, and H.~Li.
\newblock Soft rasterizer: {A} differentiable renderer for image-based 3d
  reasoning.
\newblock \emph{arXiv}, abs/1904.01786, 2019.

\bibitem[Liu et~al.(2015)Liu, Luo, Wang, and Tang]{liu2015celeba}
Z.~Liu, P.~Luo, X.~Wang, and X.~Tang.
\newblock Deep learning face attributes in the wild.
\newblock In \emph{Proc. ICCV}, 2015.

\bibitem[Loper and Black(2014)]{Loper2014}
M.~M. Loper and M.~J. Black.
\newblock {OpenDR}: An approximate differentiable renderer.
\newblock In \emph{Proc. ECCV}, 2014.

\bibitem[Luo et~al.(2018)Luo, Ren, Lin, Pang, Sun, Li, and Lin]{Luo2018SVS}
Y.~Luo, J.~Ren, M.~Lin, J.~Pang, W.~Sun, H.~Li, and L.~Lin.
\newblock Single view stereo matching.
\newblock In \emph{Proc. CVPR}, 2018.

\bibitem[Moniz et~al.(2018)Moniz, Beckham, Rajotte, Honari, and Pal]{Moniz2018}
J.~R.~A. Moniz, C.~Beckham, S.~Rajotte, S.~Honari, and C.~Pal.
\newblock Unsupervised depth estimation, 3d face rotation and replacement.
\newblock In \emph{NeurIPS}, 2018.

\bibitem[Nguyen-Phuoc et~al.(2019)Nguyen-Phuoc, Li, Theis, Richardt, and
  Yang]{nguyen2019hologan}
T.~Nguyen-Phuoc, C.~Li, L.~Theis, C.~Richardt, and Y.-L. Yang.
\newblock Hologan: Unsupervised learning of 3d representations from natural
  images.
\newblock \emph{arXiv}, abs/1904.01326, 2019.

\bibitem[Novotny et~al.(2017)Novotny, Larlus, and Vedaldi]{Novotny17b}
D.~Novotny, D.~Larlus, and A.~Vedaldi.
\newblock Learning 3d object categories by looking around them.
\newblock In \emph{Proc. ICCV}, 2017.

\bibitem[Paysan et~al.(2009)Paysan, Knothe, Amberg, Romdhani, and
  Vetter]{bfm09}
P.~Paysan, R.~Knothe, B.~Amberg, S.~Romdhani, and T.~Vetter.
\newblock A 3d face model for pose and illumination invariant face recognition.
\newblock In \emph{The IEEE International Conference on Advanced Video and
  Signal Based Surveillance}, 2009.

\bibitem[Sahasrabudhe et~al.(2019)Sahasrabudhe, Shu, Bartrum, Guler, Samaras,
  and Kokkinos]{Sahasrabudhe2019}
M.~Sahasrabudhe, Z.~Shu, E.~Bartrum, R.~A. Guler, D.~Samaras, and I.~Kokkinos.
\newblock Lifting autoencoders: Unsupervised learning of a fully-disentangled
  3d morphable model using deep non-rigid structure from motion.
\newblock \emph{arXiv}, abs/1904.11960, 2019.

\bibitem[Sengupta et~al.(2018)Sengupta, Kanazawa, Castillo, and
  Jacobs]{Sengupta18sfsnet}
S.~Sengupta, A.~Kanazawa, C.~D. Castillo, and D.~W. Jacobs.
\newblock Sfsnet: Learning shape, refectance and illuminance of faces in the
  wild.
\newblock In \emph{Proc. CVPR}, 2018.

\bibitem[Shu et~al.(2018)Shu, Sahasrabudhe, Guler, Samaras, Paragios, and
  Kokkinos]{Shu2018}
Z.~Shu, M.~Sahasrabudhe, R.~A. Guler, D.~Samaras, N.~Paragios, and I.~Kokkinos.
\newblock Deforming autoencoders: Unsupervised disentangling of shape and
  appearance.
\newblock In \emph{Proc. ECCV}, 2018.

\bibitem[Simonyan and Zisserman(2015)]{Simonyan15}
K.~Simonyan and A.~Zisserman.
\newblock Very deep convolutional networks for large-scale image recognition.
\newblock In \emph{International Conference on Learning Representations}, 2015.

\bibitem[Suwajanakorn et~al.(2018)Suwajanakorn, Snavely, Tompson, and
  Norouzi]{Suwajanakorn2018}
S.~Suwajanakorn, N.~Snavely, J.~Tompson, and M.~Norouzi.
\newblock Discovery of latent 3d keypoints via end-to-end geometric reasoning.
\newblock In \emph{NeurIPS}, 2018.

\bibitem[Szab{\'{o}} and Favaro(2018)]{Szabo18}
A.~Szab{\'{o}} and P.~Favaro.
\newblock Unsupervised 3d shape learning from image collections in the wild.
\newblock \emph{arXiv}, abs/1811.10519, 2018.

\bibitem[Tewari et~al.(2017)Tewari, Zoll{\"o}fer, Kim, Garrido, Bernard, Perez,
  and C.]{tewari17MoFA}
A.~Tewari, M.~Zoll{\"o}fer, H.~Kim, P.~Garrido, F.~Bernard, P.~Perez, and
  Theobalt C.
\newblock Mofa: Model-based deep convolutional face autoencoder for
  unsupervised monocular reconstruction.
\newblock In \emph{Proc. ICCV}, 2017.

\bibitem[Thewlis et~al.(2017{\natexlab{a}})Thewlis, Bilen, and
  Vedaldi]{Thewlis17}
J.~Thewlis, H.~Bilen, and A.~Vedaldi.
\newblock Unsupervised learning of object landmarks by factorized spatial
  embeddings.
\newblock In \emph{Proc. ICCV}, 2017{\natexlab{a}}.

\bibitem[Thewlis et~al.(2017{\natexlab{b}})Thewlis, Bilen, and
  Vedaldi]{Thewlis17b}
J.~Thewlis, H.~Bilen, and A.~Vedaldi.
\newblock Unsupervised learning of object frames by dense equivariant image
  labelling.
\newblock In \emph{NeurIPS}, 2017{\natexlab{b}}.

\bibitem[Thewlis et~al.(2018)Thewlis, Bilen, and Vedaldi]{Thewlis18}
J.~Thewlis, H.~Bilen, and A.~Vedaldi.
\newblock Modelling and unsupervised learning of symmetric deformable object
  categories.
\newblock In \emph{NeurIPS}, 2018.

\bibitem[Tung et~al.(2017)Tung, Harley, Seto, and Fragkiadaki]{tung2017}
H.-Y.~F. Tung, A.~W. Harley, W.~Seto, and K.~Fragkiadaki.
\newblock Adversarial inverse graphics networks: {L}earning 2d-to-3d lifting
  and image-to-image translation from unpaired supervision.
\newblock In \emph{Proc. ICCV}, 2017.

\bibitem[Ummenhofer et~al.(2017)Ummenhofer, Zhou, Uhrig, Mayer, Ilg,
  Dosovitskiy, and Brox]{ummenhofer2017demon}
B.~Ummenhofer, H.~Zhou, J.~Uhrig, N.~Mayer, E.~Ilg, A.~Dosovitskiy, and
  T.~Brox.
\newblock Demon: Depth and motion network for learning monocular stereo.
\newblock In \emph{Proc. CVPR}, pages 5038--5047, 2017.

\bibitem[Wang et~al.(2018)Wang, Miguel~Buenaposada, Zhu, and
  Lucey]{Wang_2018_CVPR}
C.~Wang, J.~Miguel~Buenaposada, R.~Zhu, and S.~Lucey.
\newblock Learning depth from monocular videos using direct methods.
\newblock In \emph{Proc. CVPR}, 2018.

\bibitem[Wang et~al.(2017)Wang, Shu, Cheng, Panagakis, Samaras, and
  Zafeiriou]{wang2017adversarial}
M.~Wang, Z.~Shu, S.~Cheng, Y.~Panagakis, D.~Samaras, and S.~Zafeiriou.
\newblock An adversarial neuro-tensorial approach for learning disentangled
  representations.
\newblock \emph{IJCV}, pages 1--20, 2017.

\bibitem[Woodham(1980)]{woodham1980photometric}
R.~J. Woodham.
\newblock Photometric method for determining surface orientation from multiple
  images.
\newblock \emph{Optical engineering}, 19\penalty0 (1):\penalty0 191139, 1980.

\bibitem[Xiao et~al.(2010)Xiao, Hays, Ehinger, Oliva, and
  Torralba]{Xiao2010SUN}
J.~Xiao, J.~Hays, K.~A. Ehinger, A.~Oliva, and A.~Torralba.
\newblock Sun database: Large-scale scene recognition from abbey to zoo.
\newblock In \emph{Proc. CVPR}, 2010.

\bibitem[Yin et~al.(2008)Yin, Chen, Sun, Worm, and Reale]{Yin2008}
L.~Yin, X.~Chen, Y.~Sun, T.~Worm, and M.~Reale.
\newblock A high-resolution 3d dynamic facial expression database.
\newblock In \emph{Proc. Int. Conf. Autom. Face and Gesture Recog.}, 2008.

\bibitem[Zhang et~al.(2008)Zhang, Sun, and Tang]{zhang2008cat}
W.~Zhang, J.~Sun, and X.~Tang.
\newblock Cat head detection - how to effectively exploit shape and texture
  features.
\newblock In \emph{Proc. ECCV}, 2008.

\bibitem[Zhang et~al.(2014)Zhang, Yin, Cohn, Canavan, Reale, Horowitz, Liu, and
  Girard]{ZHANG2014692}
X.~Zhang, L.~Yin, J.~F. Cohn, S.~Canavan, M.~Reale, A.~Horowitz, P.~Liu, and
  J.~M. Girard.
\newblock Bp4d-spontaneous: a high-resolution spontaneous 3d dynamic facial
  expression database.
\newblock \emph{Image and Vision Computing}, 32\penalty0 (10):\penalty0
  692--706, 2014.

\bibitem[Zhou et~al.(2017)Zhou, Brown, Snavely, and Lowe]{zhou2017unsupervised}
T.~Zhou, M.~Brown, N.~Snavely, and D.~G. Lowe.
\newblock Unsupervised learning of depth and ego-motion from video.
\newblock In \emph{Proc. CVPR}, 2017.

\bibitem[Zhu et~al.(2018)Zhu, Zhang, Zhang, Wu, Torralba, Tenenbaum, and
  Freeman]{zhu2018von}
J.-Y. Zhu, Z.~Zhang, C.~Zhang, J.~Wu, A.~Torralba, J.~B. Tenenbaum, and W.~T.
  Freeman.
\newblock Visual object networks: Image generation with disentangled 3{D}
  representations.
\newblock In \emph{NeurIPS}, 2018.

\end{thebibliography}
\bibliographystyle{plainnat}
\clearpage\newpage\section{Appendix}

\subsection{Further Implementation Details}
We will release the code and the datasets for future benchmarks upon acceptance of this paper. 

Table~\ref{tab:dataset} summarizes the number of images in each of datasets used in this paper. We use an image size of $64\times 64$ in all experiments. 
We also report all hyper-parameter settings in \cref{tab:param_set}. Our models were trained for around $35$k iterations (e.g., $30$ epochs for the synthetic face dataset), which translates to roughly one day on a Titan X Pascal GPU. 
To avoid border issues after the viewpoint transformation, we predict depth maps twice as large and crop the center after warping. 

\paragraph{Architecture}
We use standard encoder networks for both viewpoint and lighting prediction, and encoder decoder networks for albedo and depth prediction. The architecture for each network is detailed in \cref{tab:arch_depth}, \cref{tab:arch_alb}, and \cref{tab:arch_view}. Abbreviations of building blocks are defined are as follows: 

\begin{itemize}
	\item $\text{Conv}(c_{in}, c_{out}, k, s, p)$: convolution with $c_{in}$ input channels, $c_{out}$ output channels, kernel size $k$, stride $s$ and padding $p$.
	\item $\text{Deconv}(c_{in}, c_{out}, k, s, p)$: convolution with $c_{in}$ input channels, $c_{out}$ output channels, kernel size $k$, stride $s$ and padding $p$.
	\item $\text{BN}$: batch normalization.
	\item $\text{DBE}(c_{in}, k)$: dense encoder block with $k$ $3 \times 3$ convolutions with $c_{in}$ channels, each followed by batch normalization and ReLU.
	\item $\text{TBE}(c_{in}, c_{out}, s)$: encoder transition block with $4 \times 4$ convolutions with $c_{in}$ input channels, $c_{out}$ output channels and stride $s$, each followed by batch normalization and LeakyReLU.
	\item $\text{DBD}(c_{in}, k)$: dense decoder block with $k$ $3 \times 3$ deconvolutions with $c_{in}$ channels, each followed by batch normalization and ReLU.
	\item $\text{TBD}(c_{in}, c_{out}, s)$: encoder transition block with $4 \times 4$ deconvolutions with $c_{in}$ input channels, $c_{out}$ output channels and stride $s$, each followed by batch normalization and ReLU.
\end{itemize}


\begin{table}[h]
\footnotesize
\caption{Dataset split sizes for training, validation and testing.}
\label{tab:dataset}
\begin{center}
\begin{tabular}{lrrrr}
\toprule
            & Total     & Train     & Val       & Test      \\ \midrule
 Syn Face   & $187,500$ & $150,000$ & $18,750$  & $18,750$  \\
 CelebA     & $202,598$ & $162,769$ & $19,867$  & $19,962$  \\
 3DFAW      & $23,308$  & $13,671$  & $4,725$   & $4,912$   \\
 Cats       & $9,997$   & $7,997$   & $1,000$   & $1,000$   \\
 Cars       & $35,140$  & $24,580$  & $3,520$   & $7,040$   \\
\bottomrule
\end{tabular}
\end{center}
\end{table}

\begin{table}[h]
\begin{floatrow}[2]
  
  \ttabbox{
    \footnotesize
    \begin{tabular}{l}
        \toprule
         Encoder \\ \midrule
         Conv(3, 64, 4, 2, 1) + LeakyReLU(0.2) \\
         Conv(64, 128, 4, 2, 1) + BN + LeakyReLU(0.2) \\
         Conv(128, 256, 4, 2, 1) + BN + LeakyReLU(0.2) \\
         Conv(256, 512, 4, 2, 1) + BN + LeakyReLU(0.2) \\
         Conv(512, 512, 4, 2, 1) + BN + LeakyReLU(0.2) \\
         Conv(512, 128, 2, 1, 0) + BN + LeakyReLU(0.2) \\ \midrule
         Decoder \\ \midrule
         Deconv(128, 512, 2, 1, 0) + BN + ReLU \\
         Deconv(512, 512, 4, 2, 1) + BN + ReLU \\
         Deconv(512, 256, 4, 2, 1) + BN + ReLU \\
         Deconv(256, 128, 4, 2, 1) + BN + ReLU \\
         Deconv(128, 64, 4, 2, 1) + BN + ReLU \\
         Deconv(64, 64, 4, 2, 1) + BN + ReLU \\
         Conv(64, 1, 5, 1, 2) + Tanh \\
        \bottomrule
        \end{tabular}
    }{
    \caption{Depth network}
    \label{tab:arch_depth}
    }
    
    \ttabbox{
    \footnotesize
    \begin{tabular}{ll}
        \toprule
         Dense Encoder \\ \midrule
         Conv(3, 64, 4, 2, 1) \\
         DBE(64, 6) + TBE(64, 128, 2) \\
         DBE(128, 12) + TBE(128, 256, 2) \\
         DBE(256, 24) + TBE(256, 512, 2) \\
         DBE(512, 16) + TBE(512, 128, 4) \\
         Sigmoid \\ \midrule
         Dense Decoder \\ \midrule
         Deconv(128, 512, 4, 1, 0) \\
         DBD(512, 16) + TBD(512, 256, 2) \\
         DBD(256, 24) + TBD(256, 128, 2) \\
         DBD(128, 12) + TBD(128, 64, 2) \\
         DBD(64, 6) + TBD(64, 64, 2) \\
         BN + ReLU + Conv(64, 3, 5, 1, 2) \\
         Tanh\\
        \bottomrule
        \end{tabular}
  }{
    \caption{Albedo network}
    \label{tab:arch_alb}
  }
\end{floatrow}
\end{table}

\begin{table}[h]
\begin{floatrow}[2]
  
    \ttabbox{
    \footnotesize
        \begin{tabular}{lr}
        \toprule
         Parameter & Value/Range \\ \midrule
         Optimizer & Adam \\
         Learning rate & $1\times 10^{-4}$ \\
         Number of epochs & $30$ \\
         Batch size & $128$ \\
         Loss weight $\lambda_1$ & $1$ \\
         Loss weight $\lambda_{\text{perc}}$ & $0.003$ \\
         Loss weight $\lambda_{\text{d}}$ & $0.05$ \\
         Loss weight $\lambda_{\text{vp}}$ & $1$ \\ \midrule
         Depth (Human face) & $(0.4,0.6)$ \\
         Depth (Cat head) & $(0.4,0.6)$ \\
         Depth (Car) & $(28.5,31.5)$ \\
         Albedo & $(0,1)$ \\
         Light coefficient $k_s$ & $(0,1)$ \\
         Light coefficient $k_d$ & $(0,1)$ \\
         Light direction x/y & $(-60^\circ, 60^\circ)$ \\
         Viewpoint rotation x/y/z & $(-60^\circ, 60^\circ)$ \\
         Viewpoint translation x/y/z & $(-0.1, 0.1)$ \\
        \bottomrule
        \end{tabular}
        \vspace{6cm}
  }{
    \caption{Hyper-parameter settings}
    \label{tab:param_set}
  }
  
  \ttabbox{
    \footnotesize
        \begin{tabular}{p{16.5em}}
        \toprule
         Encoder                        \\ \midrule
         Conv(3, 32, 4, 2, 1) + ReLU    \\
         Conv(32, 64, 4, 2, 1) + ReLU   \\
         Conv(64, 128, 4, 2, 1) + ReLU  \\
         Conv(128, 256, 4, 1, 0) + ReLU \\
         Conv(256, 256, 4, 2, 1) + ReLU \\
         FC(256, output dim) + Tanh        \\
        \bottomrule
        \end{tabular}
    }{
        \caption{Viewpoint \& light networks}
        \label{tab:arch_view}
    }
    
\end{floatrow}
\end{table}

\clearpage
\subsection{More Qualitative Results}

\begin{figure}[h]
\begin{subfigure}{.09\textwidth}
  \includegraphics[width=.99\linewidth]{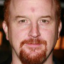}
\end{subfigure}
\hfill
\begin{subfigure}{.8\textwidth}
  \includegraphics[width=.99\linewidth]{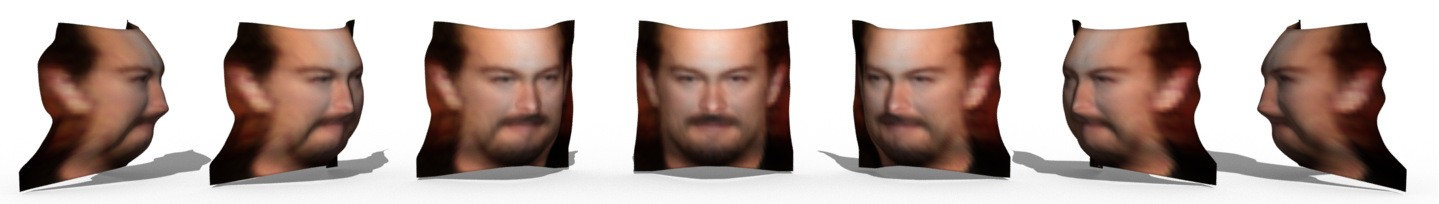}
\end{subfigure} \\ 

\begin{subfigure}{.09\textwidth}
  \includegraphics[width=.99\linewidth]{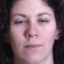}
\end{subfigure}
\hfill
\begin{subfigure}{.8\textwidth}
  \includegraphics[width=.99\linewidth]{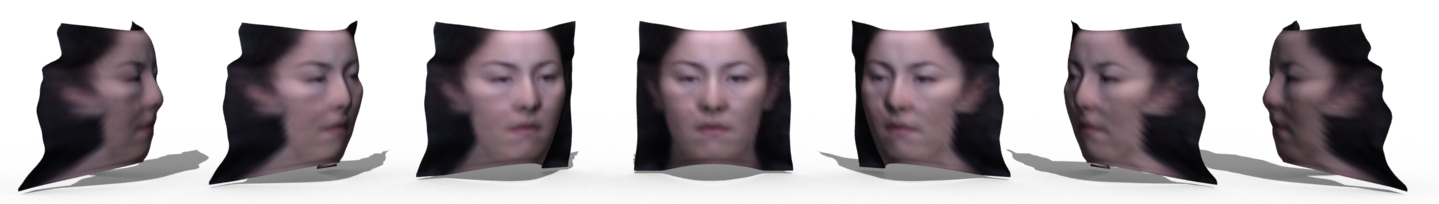}
\end{subfigure} \\

\begin{subfigure}{.09\textwidth}
  \includegraphics[width=.99\linewidth]{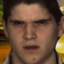}
\end{subfigure}
\hfill
\begin{subfigure}{.8\textwidth}
  \includegraphics[width=.99\linewidth]{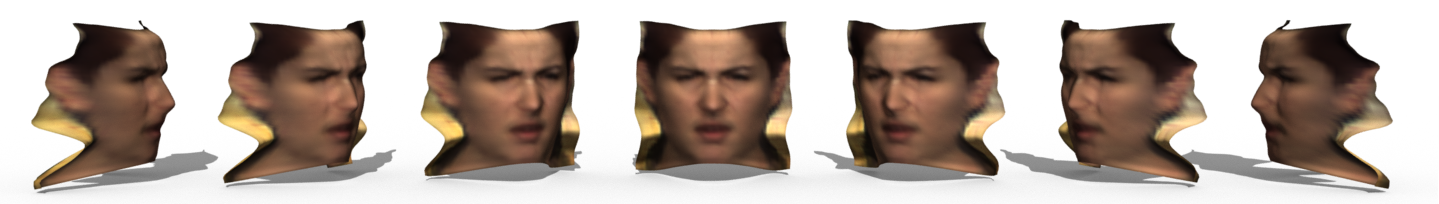}
\end{subfigure} \\

\begin{subfigure}{.09\textwidth}
  \includegraphics[width=.99\linewidth]{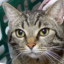}
\end{subfigure}
\hfill
\begin{subfigure}{.8\textwidth}
  \includegraphics[width=.99\linewidth]{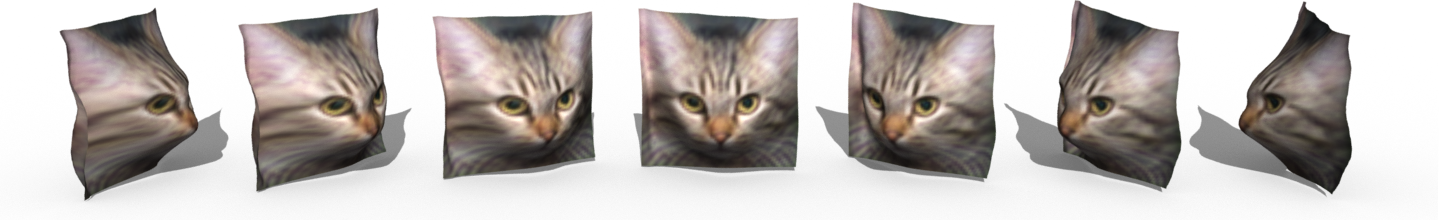}
\end{subfigure} \\

\begin{subfigure}{.09\textwidth}
  \includegraphics[width=.99\linewidth]{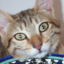}
\end{subfigure}
\hfill
\begin{subfigure}{.8\textwidth}
  \includegraphics[width=.99\linewidth]{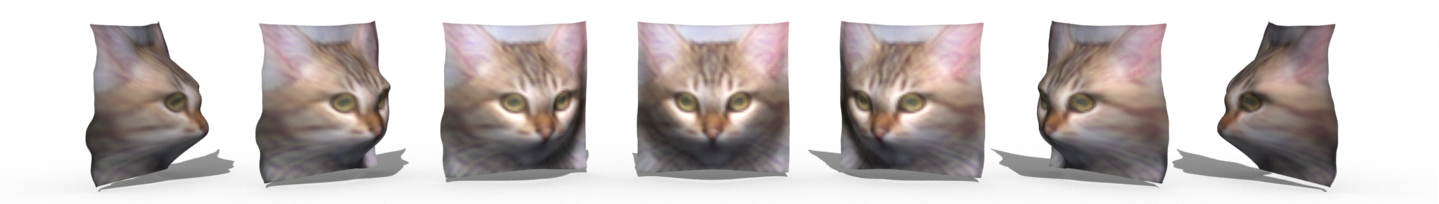}
\end{subfigure} \\

\begin{subfigure}{.09\textwidth}
  \includegraphics[width=.99\linewidth]{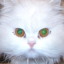}
\end{subfigure}
\hfill
\begin{subfigure}{.8\textwidth}
  \includegraphics[width=.99\linewidth]{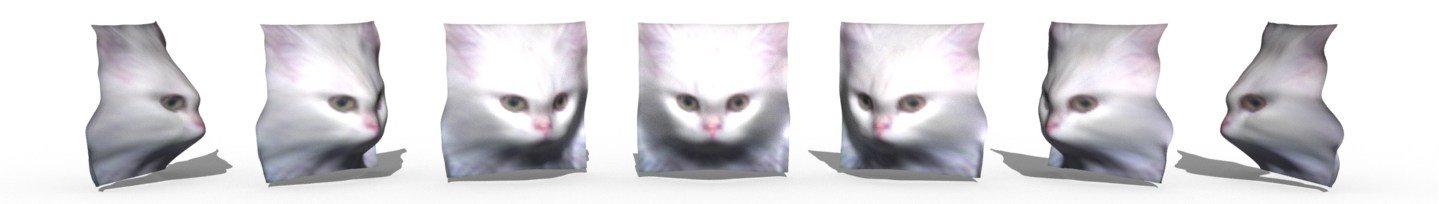}
\end{subfigure} \\

\begin{subfigure}{.09\textwidth}
  \includegraphics[width=.99\linewidth]{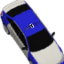}
\end{subfigure}
\hfill
\begin{subfigure}{.8\textwidth}
  \includegraphics[width=.99\linewidth]{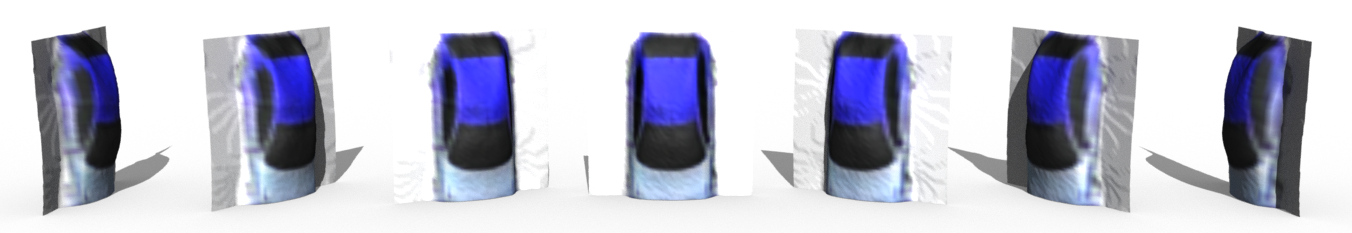}
\end{subfigure} \\

\begin{subfigure}{.09\textwidth}
  \includegraphics[width=.99\linewidth]{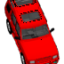}
\end{subfigure}
\hfill
\begin{subfigure}{.8\textwidth}
  \includegraphics[width=.99\linewidth]{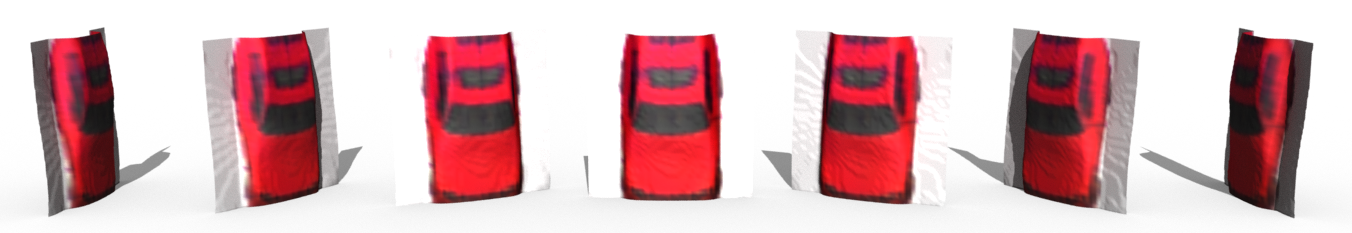}
\end{subfigure} \\

\begin{subfigure}{.09\textwidth}
  \includegraphics[width=.99\linewidth]{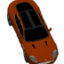}
\end{subfigure}
\hfill
\begin{subfigure}{.8\textwidth}
  \includegraphics[width=.99\linewidth]{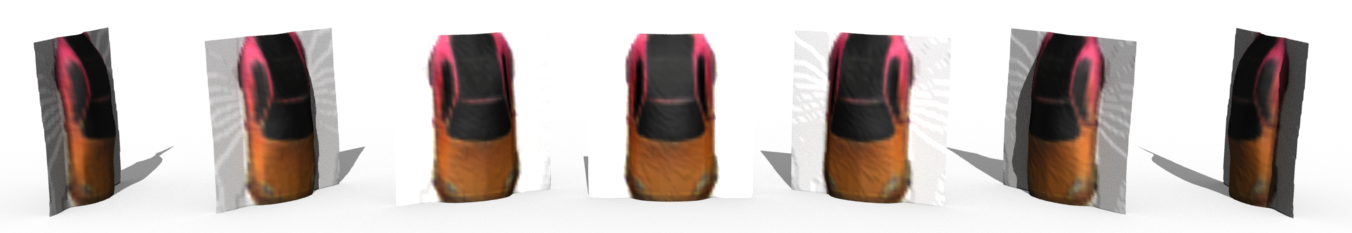}
\end{subfigure} 

\caption{More results on faces, cats and cars. The lefthand side is input, and the righthand side shows the recovered 3D object rendered from different viewpoints. }
\end{figure}

\begin{figure}[h]
	\begin{center}
	
	\begin{tabular}{ccccccc}
	
		\includegraphics[width=1.3cm]{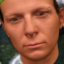} &
		\includegraphics[width=1.3cm]{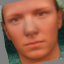} &
		\includegraphics[width=1.3cm]{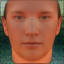} &
		\includegraphics[width=1.3cm]{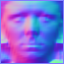} &
		\includegraphics[width=1.3cm]{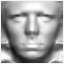} &
		\includegraphics[width=1.3cm]{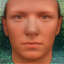} &
		\includegraphics[width=1.3cm]{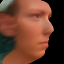} \\
		
		\includegraphics[width=1.3cm]{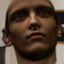} &
		\includegraphics[width=1.3cm]{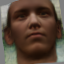} &
		\includegraphics[width=1.3cm]{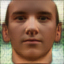} &
		\includegraphics[width=1.3cm]{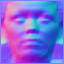} &
		\includegraphics[width=1.3cm]{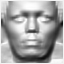} &
		\includegraphics[width=1.3cm]{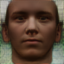} &
		\includegraphics[width=1.3cm]{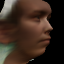} \\
		
		\includegraphics[width=1.3cm]{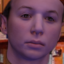} &
		\includegraphics[width=1.3cm]{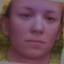} &
		\includegraphics[width=1.3cm]{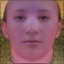} &
		\includegraphics[width=1.3cm]{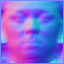} &
		\includegraphics[width=1.3cm]{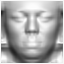} &
		\includegraphics[width=1.3cm]{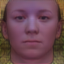} &
		\includegraphics[width=1.3cm]{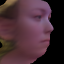} \\
		
		\includegraphics[width=1.3cm]{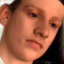} &
		\includegraphics[width=1.3cm]{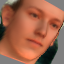} &
		\includegraphics[width=1.3cm]{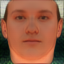} &
		\includegraphics[width=1.3cm]{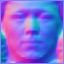} &
		\includegraphics[width=1.3cm]{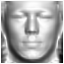} &
		\includegraphics[width=1.3cm]{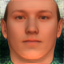} &
		\includegraphics[width=1.3cm]{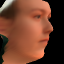} \\
		
		\includegraphics[width=1.3cm]{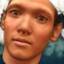} &
		\includegraphics[width=1.3cm]{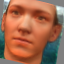} &
		\includegraphics[width=1.3cm]{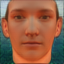} &
		\includegraphics[width=1.3cm]{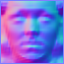} &
		\includegraphics[width=1.3cm]{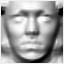} &
		\includegraphics[width=1.3cm]{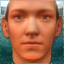} &
		\includegraphics[width=1.3cm]{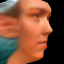} \\
		
		input & recon & albedo & normal & shading & shaded & side
	\end{tabular}
    
	\end{center}
	\caption{More results of intrinsic image decomposition on synthetic faces. }
\end{figure}

\begin{figure}[h]
	\begin{center}
	
	\begin{tabular}{ccccccc}
		\includegraphics[width=1.3cm]{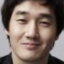} &
		\includegraphics[width=1.3cm]{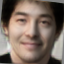} &
		\includegraphics[width=1.3cm]{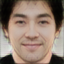} &
		\includegraphics[width=1.3cm]{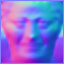} &
		\includegraphics[width=1.3cm]{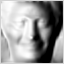} &
		\includegraphics[width=1.3cm]{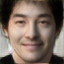} &
		\includegraphics[width=1.3cm]{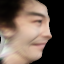} \\
		
		\includegraphics[width=1.3cm]{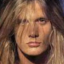} &
		\includegraphics[width=1.3cm]{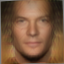} &
		\includegraphics[width=1.3cm]{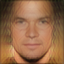} &
		\includegraphics[width=1.3cm]{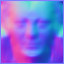} &
		\includegraphics[width=1.3cm]{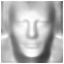} &
		\includegraphics[width=1.3cm]{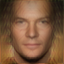} &
		\includegraphics[width=1.3cm]{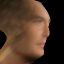} \\
		
		\includegraphics[width=1.3cm]{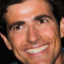} &
		\includegraphics[width=1.3cm]{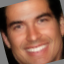} &
		\includegraphics[width=1.3cm]{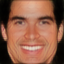} &
		\includegraphics[width=1.3cm]{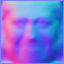} &
		\includegraphics[width=1.3cm]{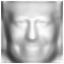} &
		\includegraphics[width=1.3cm]{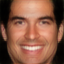} &
		\includegraphics[width=1.3cm]{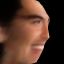} \\
		
		\includegraphics[width=1.3cm]{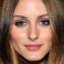} &
		\includegraphics[width=1.3cm]{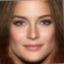} &
		\includegraphics[width=1.3cm]{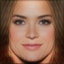} &
		\includegraphics[width=1.3cm]{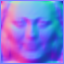} &
		\includegraphics[width=1.3cm]{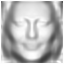} &
		\includegraphics[width=1.3cm]{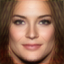} &
		\includegraphics[width=1.3cm]{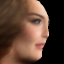} \\
		
		\includegraphics[width=1.3cm]{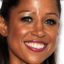} &
		\includegraphics[width=1.3cm]{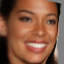} &
		\includegraphics[width=1.3cm]{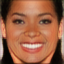} &
		\includegraphics[width=1.3cm]{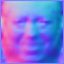} &
		\includegraphics[width=1.3cm]{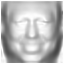} &
		\includegraphics[width=1.3cm]{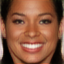} &
		\includegraphics[width=1.3cm]{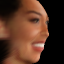} \\
		
		input & recon & albedo & normal & shading & shaded & side
	\end{tabular}
    
	\end{center}
	\caption{More results of intrinsic image decomposition on CelebA faces. }
\end{figure}

\begin{figure}[t]
	\begin{subfigure}{.1\textwidth}
      \includegraphics[width=1.3cm]{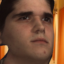}
    \end{subfigure}
    \hfill
    \begin{subfigure}{.86\textwidth}
		\begin{tabular}{ccccccc}
			\includegraphics[width=1.3cm]{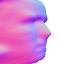} &
			\includegraphics[width=1.3cm]{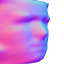} &
			\includegraphics[width=1.3cm]{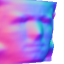} &
			\includegraphics[width=1.3cm]{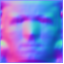} &
			\includegraphics[width=1.3cm]{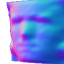} &
			\includegraphics[width=1.3cm]{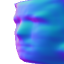} &
			\includegraphics[width=1.3cm]{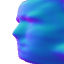} \\
			\includegraphics[width=1.3cm]{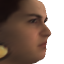} &
			\includegraphics[width=1.3cm]{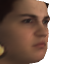} &
			\includegraphics[width=1.3cm]{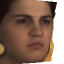} &
			\includegraphics[width=1.3cm]{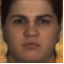} &
			\includegraphics[width=1.3cm]{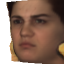} &
			\includegraphics[width=1.3cm]{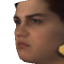} &
			\includegraphics[width=1.3cm]{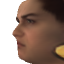} 
		\end{tabular}
    \end{subfigure} \\
    
	\begin{subfigure}{.1\textwidth}
      \includegraphics[width=1.3cm]{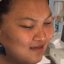}
    \end{subfigure}
    \hfill
    \begin{subfigure}{.86\textwidth}
		\begin{tabular}{ccccccc}
			\includegraphics[width=1.3cm]{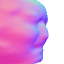} &
			\includegraphics[width=1.3cm]{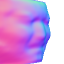} &
			\includegraphics[width=1.3cm]{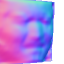} &
			\includegraphics[width=1.3cm]{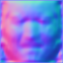} &
			\includegraphics[width=1.3cm]{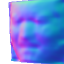} &
			\includegraphics[width=1.3cm]{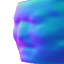} &
			\includegraphics[width=1.3cm]{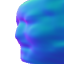} \\
			\includegraphics[width=1.3cm]{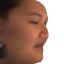} &
			\includegraphics[width=1.3cm]{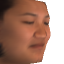} &
			\includegraphics[width=1.3cm]{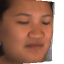} &
			\includegraphics[width=1.3cm]{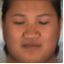} &
			\includegraphics[width=1.3cm]{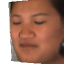} &
			\includegraphics[width=1.3cm]{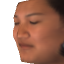} &
			\includegraphics[width=1.3cm]{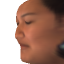} 
		\end{tabular}
    \end{subfigure}
    
        
	\begin{subfigure}{.1\textwidth}
      \includegraphics[width=1.3cm]{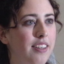}
    \end{subfigure}
    \hfill
    \begin{subfigure}{.86\textwidth}
		\begin{tabular}{ccccccc}
			\includegraphics[width=1.3cm]{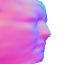} &
			\includegraphics[width=1.3cm]{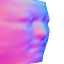} &
			\includegraphics[width=1.3cm]{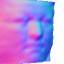} &
			\includegraphics[width=1.3cm]{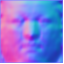} &
			\includegraphics[width=1.3cm]{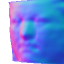} &
			\includegraphics[width=1.3cm]{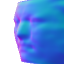} &
			\includegraphics[width=1.3cm]{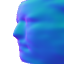} \\
			\includegraphics[width=1.3cm]{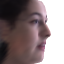} &
			\includegraphics[width=1.3cm]{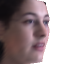} &
			\includegraphics[width=1.3cm]{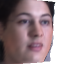} &
			\includegraphics[width=1.3cm]{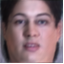} &
			\includegraphics[width=1.3cm]{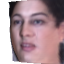} &
			\includegraphics[width=1.3cm]{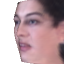} &
			\includegraphics[width=1.3cm]{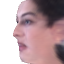} 
		\end{tabular}
    \end{subfigure} \\
    
	\begin{subfigure}{.1\textwidth}
      \includegraphics[width=1.3cm]{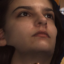}
    \end{subfigure}
    \hfill
    \begin{subfigure}{.86\textwidth}
		\begin{tabular}{ccccccc}
			\includegraphics[width=1.3cm]{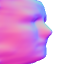} &
			\includegraphics[width=1.3cm]{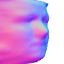} &
			\includegraphics[width=1.3cm]{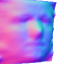} &
			\includegraphics[width=1.3cm]{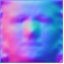} &
			\includegraphics[width=1.3cm]{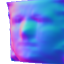} &
			\includegraphics[width=1.3cm]{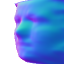} &
			\includegraphics[width=1.3cm]{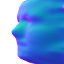} \\
			\includegraphics[width=1.3cm]{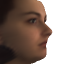} &
			\includegraphics[width=1.3cm]{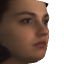} &
			\includegraphics[width=1.3cm]{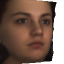} &
			\includegraphics[width=1.3cm]{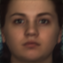} &
			\includegraphics[width=1.3cm]{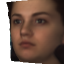} &
			\includegraphics[width=1.3cm]{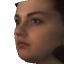} &
			\includegraphics[width=1.3cm]{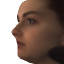} 
		\end{tabular}
    \end{subfigure} \\
    
	\begin{subfigure}{.1\textwidth}
      \includegraphics[width=1.3cm]{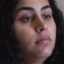}
    \end{subfigure}
    \hfill
    \begin{subfigure}{.86\textwidth}
		\begin{tabular}{ccccccc}
			\includegraphics[width=1.3cm]{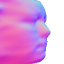} &
			\includegraphics[width=1.3cm]{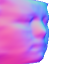} &
			\includegraphics[width=1.3cm]{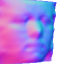} &
			\includegraphics[width=1.3cm]{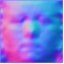} &
			\includegraphics[width=1.3cm]{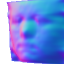} &
			\includegraphics[width=1.3cm]{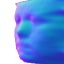} &
			\includegraphics[width=1.3cm]{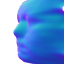} \\
			\includegraphics[width=1.3cm]{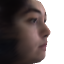} &
			\includegraphics[width=1.3cm]{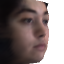} &
			\includegraphics[width=1.3cm]{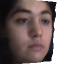} &
			\includegraphics[width=1.3cm]{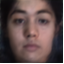} &
			\includegraphics[width=1.3cm]{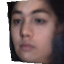} &
			\includegraphics[width=1.3cm]{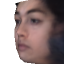} &
			\includegraphics[width=1.3cm]{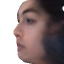} 
		\end{tabular}
    \end{subfigure} \\
    
	\caption{More results on real faces from 3DFAW. The first column is the input image, and the rest illustrates the recovered 3D face and normal map from multiple viewpoints. }
\end{figure}

\begin{figure}[t]
	\begin{center}
	
	\begin{subfigure}{.1\textwidth}
      \includegraphics[width=1.3cm]{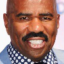}
    \end{subfigure}
    \hfill
    \begin{subfigure}{.86\textwidth}
		\begin{tabular}{ccccccc}
			\includegraphics[width=1.3cm]{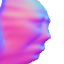} &
			\includegraphics[width=1.3cm]{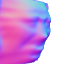} &
			\includegraphics[width=1.3cm]{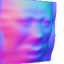} &
			\includegraphics[width=1.3cm]{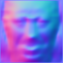} &
			\includegraphics[width=1.3cm]{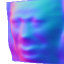} &
			\includegraphics[width=1.3cm]{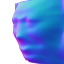} &
			\includegraphics[width=1.3cm]{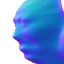} \\
			
			\includegraphics[width=1.3cm]{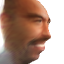} &
			\includegraphics[width=1.3cm]{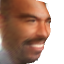} &
			\includegraphics[width=1.3cm]{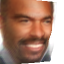} &
			\includegraphics[width=1.3cm]{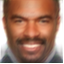} &
			\includegraphics[width=1.3cm]{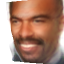} &
			\includegraphics[width=1.3cm]{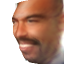} &
			\includegraphics[width=1.3cm]{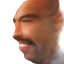} 
		\end{tabular}
    \end{subfigure} \\
    
	\begin{subfigure}{.1\textwidth}
      \includegraphics[width=1.3cm]{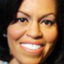}
    \end{subfigure}
    \hfill
    \begin{subfigure}{.86\textwidth}
		\begin{tabular}{ccccccc}
			\includegraphics[width=1.3cm]{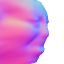} &
			\includegraphics[width=1.3cm]{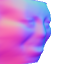} &
			\includegraphics[width=1.3cm]{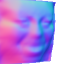} &
			\includegraphics[width=1.3cm]{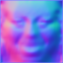} &
			\includegraphics[width=1.3cm]{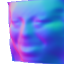} &
			\includegraphics[width=1.3cm]{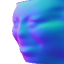} &
			\includegraphics[width=1.3cm]{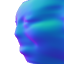} \\
			
			\includegraphics[width=1.3cm]{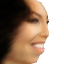} &
			\includegraphics[width=1.3cm]{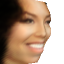} &
			\includegraphics[width=1.3cm]{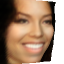} &
			\includegraphics[width=1.3cm]{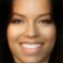} &
			\includegraphics[width=1.3cm]{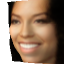} &
			\includegraphics[width=1.3cm]{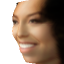} &
			\includegraphics[width=1.3cm]{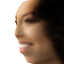} 
		\end{tabular}
    \end{subfigure} \\
    
	\begin{subfigure}{.1\textwidth}
      \includegraphics[width=1.3cm]{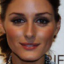}
    \end{subfigure}
    \hfill
    \begin{subfigure}{.86\textwidth}
		\begin{tabular}{ccccccc}
			\includegraphics[width=1.3cm]{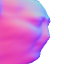} &
			\includegraphics[width=1.3cm]{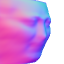} &
			\includegraphics[width=1.3cm]{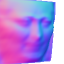} &
			\includegraphics[width=1.3cm]{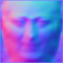} &
			\includegraphics[width=1.3cm]{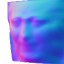} &
			\includegraphics[width=1.3cm]{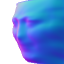} &
			\includegraphics[width=1.3cm]{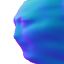} \\
			
			\includegraphics[width=1.3cm]{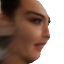} &
			\includegraphics[width=1.3cm]{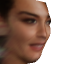} &
			\includegraphics[width=1.3cm]{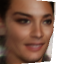} &
			\includegraphics[width=1.3cm]{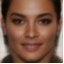} &
			\includegraphics[width=1.3cm]{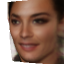} &
			\includegraphics[width=1.3cm]{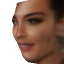} &
			\includegraphics[width=1.3cm]{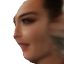} 
		\end{tabular}
    \end{subfigure} \\
    
	\begin{subfigure}{.1\textwidth}
      \includegraphics[width=1.3cm]{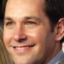}
    \end{subfigure}
    \hfill
    \begin{subfigure}{.86\textwidth}
		\begin{tabular}{ccccccc}
			\includegraphics[width=1.3cm]{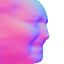} &
			\includegraphics[width=1.3cm]{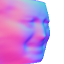} &
			\includegraphics[width=1.3cm]{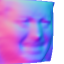} &
			\includegraphics[width=1.3cm]{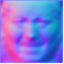} &
			\includegraphics[width=1.3cm]{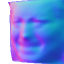} &
			\includegraphics[width=1.3cm]{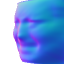} &
			\includegraphics[width=1.3cm]{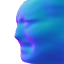} \\
			
			\includegraphics[width=1.3cm]{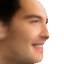} &
			\includegraphics[width=1.3cm]{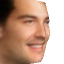} &
			\includegraphics[width=1.3cm]{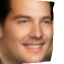} &
			\includegraphics[width=1.3cm]{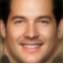} &
			\includegraphics[width=1.3cm]{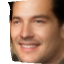} &
			\includegraphics[width=1.3cm]{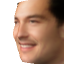} &
			\includegraphics[width=1.3cm]{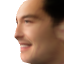}
		\end{tabular}
    \end{subfigure} \\
    
	\begin{subfigure}{.1\textwidth}
      \includegraphics[width=1.3cm]{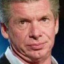}
    \end{subfigure}
    \hfill
    \begin{subfigure}{.86\textwidth}
		\begin{tabular}{ccccccc}
			\includegraphics[width=1.3cm]{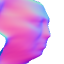} &
			\includegraphics[width=1.3cm]{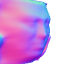} &
			\includegraphics[width=1.3cm]{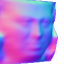} &
			\includegraphics[width=1.3cm]{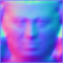} &
			\includegraphics[width=1.3cm]{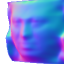} &
			\includegraphics[width=1.3cm]{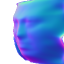} &
			\includegraphics[width=1.3cm]{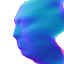} \\
			
			\includegraphics[width=1.3cm]{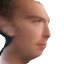} &
			\includegraphics[width=1.3cm]{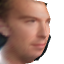} &
			\includegraphics[width=1.3cm]{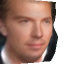} &
			\includegraphics[width=1.3cm]{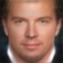} &
			\includegraphics[width=1.3cm]{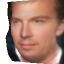} &
			\includegraphics[width=1.3cm]{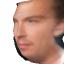} &
			\includegraphics[width=1.3cm]{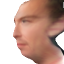}
		\end{tabular}
    \end{subfigure} \\
    
	\end{center}
	\caption{More results on real faces from CelebA. The first column is input image, and the rest illustrates the recovered 3D face and normal map from multiple viewpoints. }
\end{figure}

\end{document}